\documentclass[final]{cvpr}

\usepackage{times}
\usepackage{epsfig}
\usepackage{graphicx}
\usepackage{amsmath}
\usepackage{amssymb}

\usepackage{makecell}
\usepackage{verbatim}
\usepackage{subfigure}
\usepackage{caption}
\usepackage{longtable}
\usepackage{setspace}
\usepackage{multirow}
\usepackage{booktabs}

\usepackage[pagebackref=true,breaklinks=true,colorlinks,bookmarks=false]{hyperref}

\def\cvprPaperID{5722} 
\def\confYear{CVPR 2021}

\begin{document}

\title{Universal Face Restoration With Memorized Modulation}

\author{Jia Li, Huaibo Huang, Xiaofei Jia, Ran He\\
National Laboratory of Pattern Recognition, CASIA\\
Center for Excellence in Brain Science and Intelligence Technology, CAS\\
School of Artificial Intelligence, University of Chinese Academy of Sciences, Beijing, China\\
}

\maketitle
\begin{abstract}
  Blind face restoration (BFR) is a challenging problem because of the uncertainty of the degradation patterns. This paper proposes a Restoration with Memorized Modulation (RMM) framework for universal BFR in diverse degraded scenes and heterogeneous domains. We apply random noise as well as unsupervised wavelet memory to adaptively modulate the face-enhancement generator, considering attentional denormalization in both layer and instance levels. Specifically, in the training stage, the low-level spatial feature embedding, the wavelet memory embedding obtained by wavelet transformation of the high-resolution image, as well as the disentangled high-level noise embeddings are integrated, with the guidance of attentional maps generated from layer normalization, instance normalization and the original feature map. These three embeddings are respectively associated with the spatial content, high-frequency texture details, and a learnable universal prior against other blind image degradation patterns. We store the spatial feature of the low-resolution image and the corresponding wavelet style code as key and value in the memory unit, respectively. In the test stage, the wavelet memory value whose corresponding spatial key is the most matching with that of the inferred image is retrieved to modulate the generator. Moreover, the universal prior learned from the random noise has been memorized by training the modulation network. Experimental results show the superiority of the proposed method compared with the state-of-the-art models, and a good generalization in the wild.
\end{abstract}

\section{Introduction}

Face super-resolution (FSR) is a challenging generation task, especially when it comes to large scale factors and unpaired real scenes. Generally, the low-quality (LQ) image contains the low-level spatial content information. The high-level texture details of the high-quality (HQ) image can be formulated as the high-frequency wavelet coefficients \cite{invert, wave2018, wavelet2017}, which can potentially defend against the down-sampling degradation. Other degradation patterns in the real world are randomly comprised of Gaussian blurring, motion blurring, image noise, JPEG compaction, etc. This makes BFR more difficult than FSR. In this paper, we adversarially defend these complex degradations using noise as a universal learnable prior to modulate the blind face restoration. Comprehensively, we consider the representation learning of the degraded image, high-frequency wavelet prior, and an additional universal prior,  which is different from the recent reference-based \cite{warpnet2018, exemplar2020, pool2020, copy2020, masa2021}, HQ-dictionary-based \cite{dfdnet2020} and pretrained-model-based \cite{pulse2020, bank2021, gfp2021} SR methods that have shown great potential to improve the restoration quality.

Memory networks were firstly proposed in \cite{memory2015} for the task of question answering where the memory acts as an effective knowledge base. As for the computer vision field, there are some applications, such as image captioning \cite{m1}, text-to-image synthesis \cite{m2} and object segmentation \cite{m3}. We first propose to realize BFR with wavelet coefficient memory knowledge. 
We address SR task via attention-guided feature modulation of noise and wavelet memory, not relying on additional guidance or prior from HQ reference image \cite{warpnet2018, example2019} in the inference stage, or pretrained well-performing models (e.g. StyleGAN \cite{style2019, stylegan2_2020}) in the training stage \cite{bank2021,gfp2021}. We propose a novel BFR model called RMM that synthesizes high fidelity high-resolution (HR) face results both for the synthetic and real-world LQ images. Specifically, RMM uses the low resolution (LR) face image to carry out adaptive spatial modulation of the multi-scale decoder features, and makes use of the disentangled multi-scale noise embeddings and high-frequency wavelet code to implement the BFR information modulation. Actually, the noise embedding is resolution-independent whose latent size is fixed and with Gaussian distribution, wheres wavelet memory embedding is resolution-dependent whose distribution of style code is related to the degradation degree, i.e., the larger the downsampling scale is, the more feature dimensions wavelet embedding has.  

Adaptive feature modulation has shown great benefit to improving the desired style transferring for various synthesis tasks. \cite{spade2019} synthesizes photorealistic images using SPatially-Adaptive DEnormalization (SPADE) based on the semantic layout modulation. StyleGAN \cite{style2019, stylegan2_2020} generates high-quality images via adaptive instance normalization (AdaIN) based on noise modulation. \cite{faceshifter_2020} combines SPADE of the target face and AdaIN of the source identity to address the high fidelity face swapping. Furthermore, \cite{ugatit2020} proposes Adaptive Layer-Instance Normalization (AdaLIN) to flexibly control the fidelity of shape and texture cross the heterogeneous domains, where instance normalization (IN) \cite{in2016} and layer normalization (LN) \cite{ln2016} are combined via a learnable parameter. \cite{anigan2021} adopts adaptive point-wise layer instance normalization (AdaPoLIN) by combining IN and LN via a convolutional layer in an all-channel manner. Different from the above adaptive modulation manner, in this paper, RMM pays attention to three-level \textbf{A}ttentional maps from the \textbf{O}riginal, \textbf{L}ayer-\textbf{N}ormalized and \textbf{I}nstance-\textbf{N}ormalized features, denoted as AdaAO, AdaALN and AdaAIN. AdaAO integrates the feature maps from AdaALN and AdaAIN that adaptively fuse the LR feature, wavelet memory and noise prior with corresponding attentional maps in the layer and instance modulation view, respectively. This enables RMM to comprehensively control the structure and detail textures with consideration of the local and global statistics for the multi-level feature maps. 
\begin{figure}[t]
\begin{center}
   \includegraphics[width=8cm]{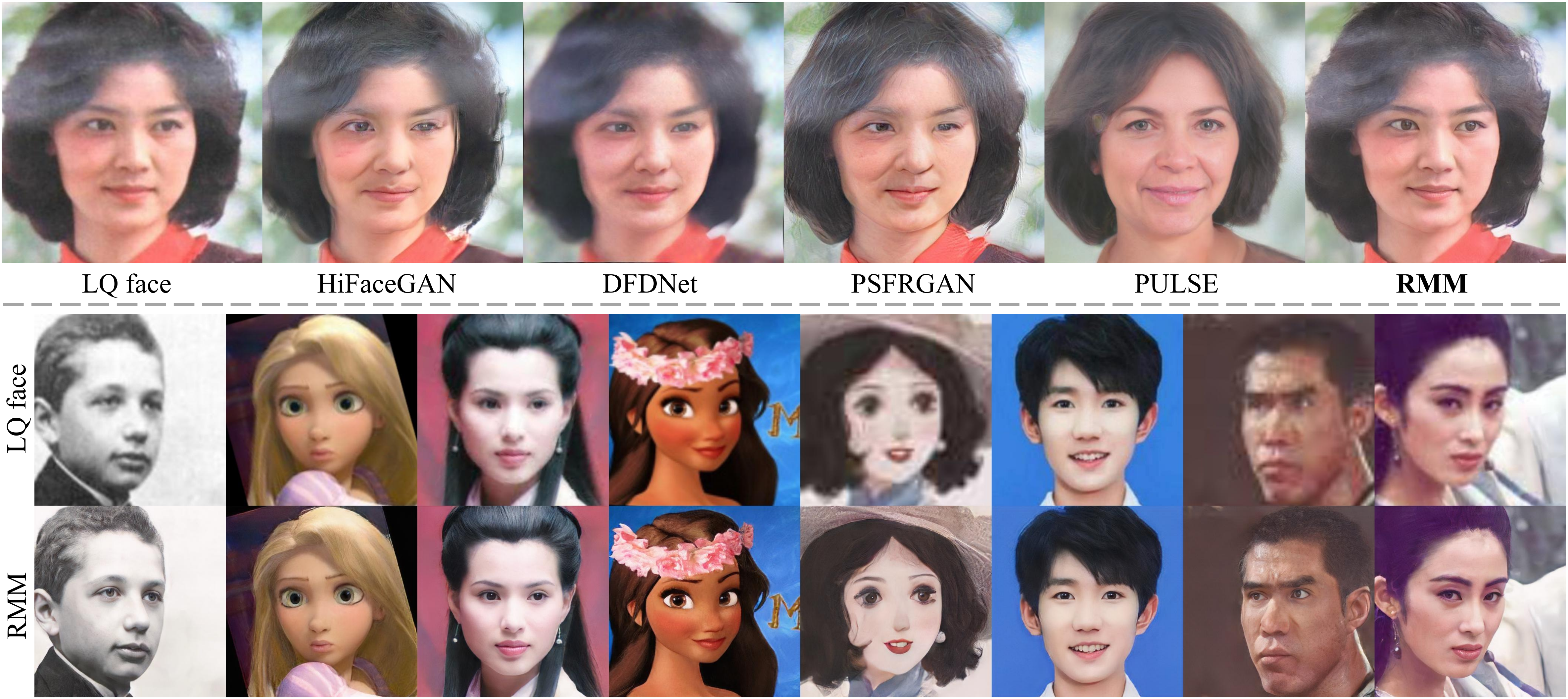}
\end{center}
\vspace{-13.pt}
   \caption{The top is the comparison with HiFaceGAN \cite{hifacegan}, DFDNet \cite{dfdnet2020}, PSFRGAN \cite{psfr}, and PULSE \cite{pulse2020}. RMM achieves higher fidelity SR results where the input is the LQ face in the real world. The bottom shows more SR results of RMM in the wild, even in the heterogeneous domains (e.g. cols 2, 4, 5). The first and second row is the LQ images and the enhanced results, respectively.}
   \vspace{-16.pt}
\label{fig:start}
\end{figure}
Our main contributions are as follows:
\begin{itemize}
\item We first study the challenging Universal Face Restoration (UFR) task and propose a novel framework RMM. By considering the learnable universal prior, and unsupervised wavelet memory, our memorized modulation achieves high-quality BFR results both for the synthetic and real-world low-quality images. 
\item We propose a neat and effective RMM Module ($RM^{3}$). Concretely, we leverage adaptive attentional maps from the original, layer-normalized and instance-normalized features, i.e., AdaAO, AdaALN and AdaAIN, to implement the quality improvement both visually and quantitatively.
\item We first propose an effective and efficient Wavelet Memory Module (WMM) that stores spatial features of the LQ images and high-frequency wavelet coefficient information, which enables to adaptively guide the high-quality blind face restoration.
\end{itemize}

\section{Related Work}
\label{gen_inst}

\paragraph{Single Image Restoration}
With the development of CNNs, GANs and flow-based model, image restoration from a single image has made great progress in many fields, including image super-resolution \cite{sr1, sr2, sr3}, deblurring \cite{deblur1, deblur2}, decompression \cite{com1, com2}, and  denoising \cite{denoise1, denoise2}. The typical SR task without guidance is WaveletSR \cite{wavelet2017} where the predicted wavelet coefficients of HQ images are composed to reconstruct the restored face. \cite{invert} adopts an
Invertible Rescaling Net (IRN) to model the dual mapping of LQ and HQ images. HiFaceGAN \cite{hifacegan} uses a collaborative suppression and replenishment (CSR) framework to achieve face renovation. 

\paragraph{Model-based Image Restoration} GLEAN \cite{bank2021} utilizes pre-trained StyleGAN to provide more detail features with the distribution of HR images. GFP-GAN \cite{gfp2021} is comprised of a degradation removal module and a pretrained StyleGAN2 \cite{stylegan2_2020} as the face prior. PULSE \cite{pulse2020} looks for the optimized latent code of StyleGAN \cite{style2019} by penalizing a down-scaling loss between the LR image and the degraded SR image.
\begin{figure*}[t]
\begin{center}
   \includegraphics[width=8cm]{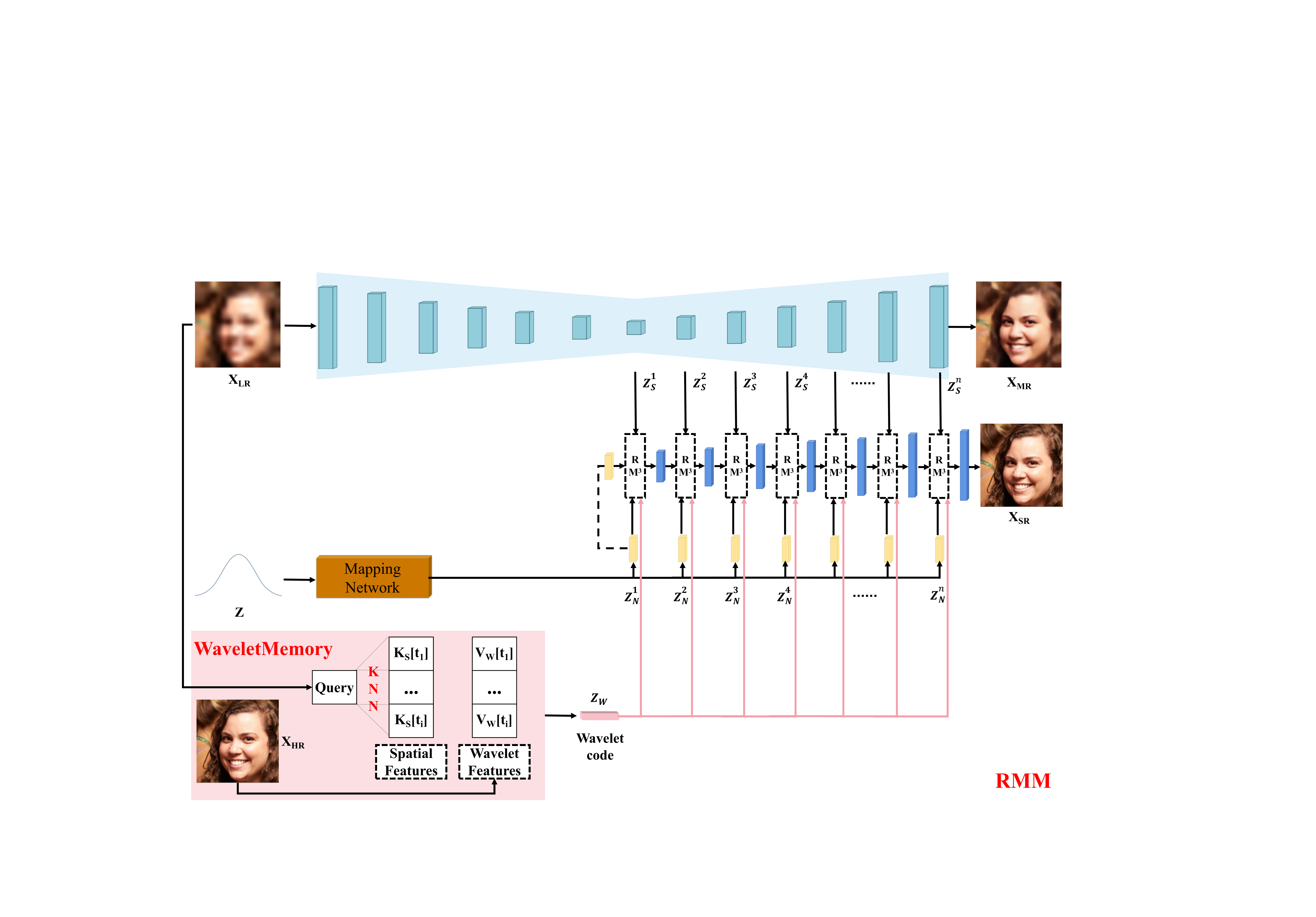}\includegraphics[width=7.5cm]{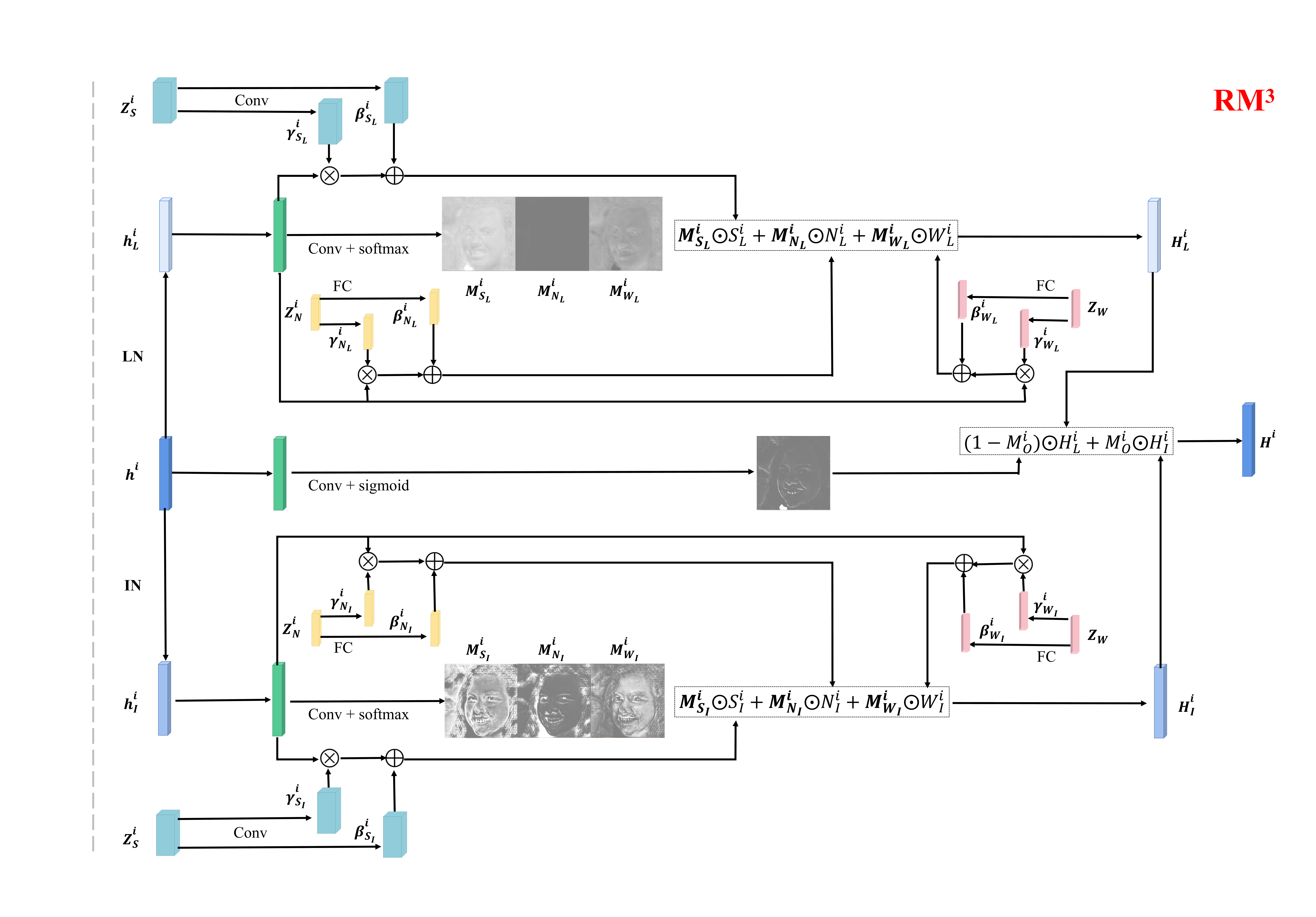}
\end{center}
\vspace{-13.pt}
   \caption{The framework of RMM. The spatial features are used to maintain the global structure during BFR, and the resolution-dependent component wavelet style code from the high-frequency wavelet coefficients controls the feature transformation of the texture details. Moreover, the resolution-independent component Gaussian noise learns a universal prior to defend against diverse real-world degradation patterns. These BFR embeddings are adaptively integrated by $RM^{3}$ module, where different kinds of attentional maps from the original, layer-normalized and instance-normalized features are taken into consideration for a comprehensive feature fusion.}
   \vspace{-16.pt}
\label{fig:pipeline}
\end{figure*}
\vspace{-13.pt}
\paragraph{Reference-based Image Restoration} WarpNet \cite{warpnet2018} makes use of the degraded and guided images to produce the restoration result. \cite{exemplar2020} incorporates prior of the reference image where both degraded, guidance features and facial landmarks are used to calculate an attention mask for adaptive high-resolution feature fusion. \cite{pool2020} builds up a universal reference pool to match the local patterns for adaptive feature enhancement of the different regions. \cite{copy2020} leverages internal and external priors to recover the HR face image. \cite{masa2021} handles different reference images by remapping the distribution of guided features to that of the LR features by spatial adaptive transferring. PSFRGAN \cite{psfr} proposes a multi-scale progressive framework for BFR based on the parsing prior. \cite{dfdnet2020} applies dictionary feature transfer (DFT) block to adaptively match with the off-line multi-scale facial component dictionaries.

\section{Approach}
\label{headings}

In this section, we will introduce our method RMM in detail, including the $RM^{3}$ module, objective function, and wavelet memory module (WMM). As shown in Figure \ref{fig:pipeline}, given a target face $X_{HR}$, we obtain the degraded image $X_{LR}$ by means of the sequential degradation methods, e.g., downsampling with the bicubic kernel followed by adding blurring and JPEG compaction. $X_{LR}$ is mapped to the refined middle-resolution face $X_{MR}$ via an Unet \cite{unet2015} with only a reconstruction constraint, and we obtain multi-scale spatial feature maps $z_{S}=\{z_{S}^{1}, z_{S}^{2}, ..., z_{S}^{n}\}$. We leverage a mapping network to disentangle the input random noise with Gaussian distribution, and utilize these prior embeddings $\{z_{N}^{1}, z_{N}^{2}, ..., z_{N}^{n}\}$ to modulate multi-scale decoder of RMM. Another branch is the wavelet style code $z_{W}$ generated from the WMM. These three embeddings are adaptively integrated by means of the $RM^{3}$ module.

\paragraph{$RM^{3}$}
For the input feature $h^{i}$ of each $RM^{3}$ block, it is modulated by three groups of affine transform parameters from $z_{S}^{i}$, $z_{N}^{i}$ and $z_{W}$, respectively. It is formulated as:
\begin{equation}
h^{i}_{I}=\frac{h^{i}-\mu^{i}_{I}}{\sigma ^{i}_{I}},
\end{equation}
\begin{equation}
\left\{S, N, W\right\}^{i}_{I}=\gamma _{\left\{S, N, W\right\}_{I}}^{i}\odot h^{i}_{I} +\beta _{\left\{S, N, W\right\}_{I}}^{i},
\end{equation}
where  $h^{i} \in \mathbb{R}^{C_{h}^{i}\times H^{i} \times W^{i}}$ is the input embedding of the current $RM^{3}$ block, $\mu^{i}_{I}$ and $\sigma^{i}_{I}$ are the means and standard deviations of $h^{i}$, and they are used to perform the instance normalization. Let $z_{S}^{i} \in \mathbb{R}^{C_{S}^{i}\times H_{S}^{i} \times W_{S}^{i}}$, $z_{W} \in \mathbb{R}^{C_{W}\times 1}$ and $z_{N}^{i} \in \mathbb{R}^{C_{N}^{i}\times 1}$ be the LR, wavelet and noise embedding, respectively. As shown in Figure \ref{fig:pipeline}, $\gamma _{S_{I}}^{i}$ and $\beta _{S_{I}}^{i}\in \mathbb{R}^{C_{h}^{i}\times H^{i}\times W^{i}}$ are obtained from $z_{S}^{i}$ using a convolutional layer. Meanwhile, $\{\gamma _{N_{I}}^{i}, \beta _{N_{I}}^{i}\in \mathbb{R}^{C_{h}^{i}\times H^{i}\times W^{i}}\}$ and $\{\gamma _{W_{I}}^{i}, \beta _{W_{I}}^{i}\in \mathbb{R}^{C_{h}^{i}\times H^{i}\times W^{i}}\}$ are obtained from $z_{N}^{i}$ and $z_{W}$ using a fully connected layer, respectively. In the AdaAIN form,  the attentional map $M^{i}_{S_{I}}$, $M^{i}_{N_{I}}$ and $M^{i}_{W_{I}}$ are obtained via a convolution and a softmax operation of $h^{i}_{I}$. The instance denormalization result $H^{i}_{I}$ is formulated as
\begin{equation}
H^{i}_{I}=S^{i}_{I}\odot M^{i}_{S_{I}}+N^{i}_{I}\odot M^{i}_{N_{I}} + W^{i}_{I}\odot M^{i}_{W_{I}},
\label{eq3}
\end{equation}
where $\odot$ means the element-wise product. Similarly, the layer normalization and denormalization are as follows
\begin{equation}
h^{i}_{L}=\frac{h^{i}-\mu^{i}_{L}}{\sigma ^{i}_{L}},
\end{equation}
\begin{equation}
\left\{S, N, W\right\}^{i}_{L}=\gamma _{\left\{S, N, W\right\}_{L}}^{i}\odot h^{i}_{L} +\beta _{\left\{S, N, W\right\}_{L}}^{i}.
\end{equation}
In the AdaALN form,  the attentional map $M^{i}_{S_{L}}$, $M^{i}_{N_{L}}$ and $M^{i}_{W_{L}}$ are obtained via a convolution and a softmax operation of $h^{i}_{L}$. The layer denormalization result $H^{i}_{L}$ is formulated as
\begin{equation}
H^{i}_{L}=S^{i}_{L}\odot M^{i}_{S_{L}}+N^{i}_{L}\odot M^{i}_{N_{L}} + W^{i}_{L}\odot M^{i}_{W_{L}}.
\label{eq3}
\end{equation}
Finally, in the AdaAO form, the attentional map $M^{i}_{O}$ is obtained via a convolution and a sigmoid operation of $h^{i}$. The $RM^{3}$ output feature $H^{i}$ is formulated as
\begin{equation}
H^{i}=H^{i}_{I}\odot M^{i}_{O}+H^{i}_{L}\odot ( 1-M^{i}_{O}).
\label{eq3}
\end{equation}
\paragraph{Objective Function} Our BFR network RMM is trained with the following constraints.

We apply the multi-scale adversarial learning \cite{p2phd}. Let $\mathcal{L}_{adv}$ be the adversarial loss to discriminate the generated BFR image $\hat{X}_{HR}$ and the real HR face $X_{HR}$ via
\begin{equation}
\mathcal{L}_{adv} = \lambda_{adv}^{(i)}\sum^{N}_{i=1}\mathbb{E}_{X}[ \log D_{i}\left( X_{HR}\right) ]+ \mathbb{E}_{Y}[ \log ( 1-D_{i}( \hat{X}_{HR}) ) ], 
\end{equation}
where $D_{i}$ means the discriminator for the downsampled image with scale factor $i$. 

We utilize two reconstruction losses used to penalize the pixel-level distance between $X_{HR}$ and $\hat{X}_{HR}$, similar to $X_{HR}$ and $X_{MR}$. They are formulated as
\begin{eqnarray}
\mathcal{L}_{rec}=\begin{cases}
\begin{array}{c}
\frac{1}{2}\triangle_{\hat{X}_{HR}}^{2}\\
\delta\cdot\left[\triangle_{\hat{X}_{HR}}-\frac{1}{2}\delta\right]
\end{array} & \begin{array}{c}
\triangle_{\hat{X}_{HR}}\leq\delta\\
otherwise
\end{array}\end{cases},
\end{eqnarray}
\begin{eqnarray}
\mathcal{L}_{rec}^{'}=\begin{cases}
\begin{array}{c}
\frac{1}{2}\triangle_{{X}_{MR}}^{2}\\
\delta\cdot\left[\triangle_{{X}_{MR}}-\frac{1}{2}\delta\right]
\end{array} & \begin{array}{c}
\triangle_{{X}_{MR}}\leq\delta\\
otherwise
\end{array}\end{cases},
\end{eqnarray}
respectively. We use Huber loss, where $\triangle_{\hat{X}_{HR}}=\left|X_{HR}-\hat{X}_{HR}\right|$, $\triangle_{{X}_{MR}}=\left|X_{HR}-{X}_{MR}\right|$, and $\delta$ is a hyperparameter.

We add the perceptual loss to improve the high-resolution blind face restoration via
\begin{equation}
\mathcal{L}_{vgg}=\lambda_{vgg}^{(i)}\sum ^{N}_{i=1} \| F_{vgg}^{(i)} (\hat{X}_{HR}) -F_{vgg}^{(i)} ({X}_{HR}) \| _{2},  
\end{equation}
where $F_{vgg}^{(i)}$ denotes the $i$-th convolution layer of the VGG19 model. We set N equal to 5. $\lambda_{vgg}^{(i)}$ is set to 1/32, 1/16, 1/8, 1/4 and 1.0 in order.



Apart from the above mentioned global objective functions, we further propose to utilize the component contextual loss to enhance the eyes and mouth areas via
\begin{equation}
\mathcal{L}_{cCX}=-\log ( CX(F_{c}^{l} (\hat{X}_{HR}),F_{c}^{l} ( X_{HR}))),
\end{equation}
where $l$ means  \emph{relu}\{3$\_$2, $4\_2$\} layers of the pretrained VGG19 network \cite{vgg2015}, and $c$ means \{\emph{left\_eye, right\_eye, mouth}\}. Contextual loss \cite{cx2018} is competent to learning the non-aligned feature matching.
Finally, the total loss of RMM is formulated as:
\begin{equation}
\mathcal{L}_{\emph{RMM}}= \mathcal{L}_{adv} +  \lambda ^{'}_{rec}\mathcal{L}_{rec}^{'} +  \lambda_{rec}\mathcal{L}_{rec} + \mathcal{L}_{vgg} +  \lambda_{cCX}\mathcal{L}_{cCX}.
\label{con:20}
\end{equation}

\paragraph{WMM} We apply an augment network Wavelet Memory Module to store the LQ spatial feature and high-frequency wavelet coefficients. As each Wavelet Memory Unit $WMU_{i}$, it contains the spatial feature key $K_{S}[t_{i}]$ and the corresponding wavelet code value $V_{W}[t_{i}]$, where $t_{i}$ means the top-$i$ value after K Nearest Neighbors operation between the key candidates and the input query. This memory mining is formulated as 
\begin{equation}
KNN(q,K_{S})=\arg \max _{i}\langle q, K_{S}\left[ i\right] \rangle,  
\end{equation}
where $\langle\cdot, \cdot\rangle$ means cosine similarity. We use pretrained ResNet18 \cite{resnet} as the spatial feature extractor, and generate the normalized query $q$ after transforming the spatial code from $pool5$ layer using a fully connected layer. 

High-frequency wavelet coefficients help to recover texture details. Consider $n$-level full wavelet packet decomposition, where $n$ means the down-scaling factor of the original image. Then the wavelet coefficient is represented as $C = (c_{1}, c_{2}, ..., c_{N_{W}})$, where $N_{W}=4^{n}$, and $c_{1}$ is the low-frequency wavelet coefficient. We treat $C_{h}\in \mathbb{R}^{(N_{W}-1)\times H^{C_{h}}\times W^{C_{h}}}$ as the high-level feature map and obtain the wavelet memory code $Z_{W}\in \mathbb{R}^{(N_{W}-1) \times1}$ by means of the Adaptive Average pooling for $C_{h}$. We use Wavelet KL divergence ($KL_{w}$) to guide the triplet loss of WMM in an unsupervised manner via
\begin{equation}
\mathcal{L}_{WMM}= max(\langle q, K_{S}[t_{p}]\rangle-\langle q, K_{S}[t_{n}]\rangle+ m, 0),
\end{equation}
where $m$ is the margin. Specifically, we minimize the distance between the positive key $K_{S}[t_{p}]$ and the query $q$ when $KL_{w}(V_{W}[t_{p}]\ ||\ Z_{W})<\eta$, while maximizing the distance between the negative key $K_{S}[t_{n}]$ and the query $q$ when $KL_{w}(V_{W}[t_{n}]\ ||\ Z_{W})>\eta$, where $\eta$ is a hyperparameter. The top-1 key $K_{S}[t_{1}]$ is updated as the mean of $q$ and the previous $\hat{K_{S}}[t_{1}]$ when $KL_{w}(V_{W}[t_{1}]\ ||\ Z_{W})<\eta$, while the new key $q$ and value $Z_{W}(q)$  will be written into wavelet memory as $K_{S}[t_{r}]$ and $V_{W}[t_{r}]$ when  $KL_{w}(V_{W}[t_{1}]\ ||\ Z_{W})>\eta$.
\begin{figure*}[ht]
\begin{center}
\includegraphics[width=0.75\linewidth]{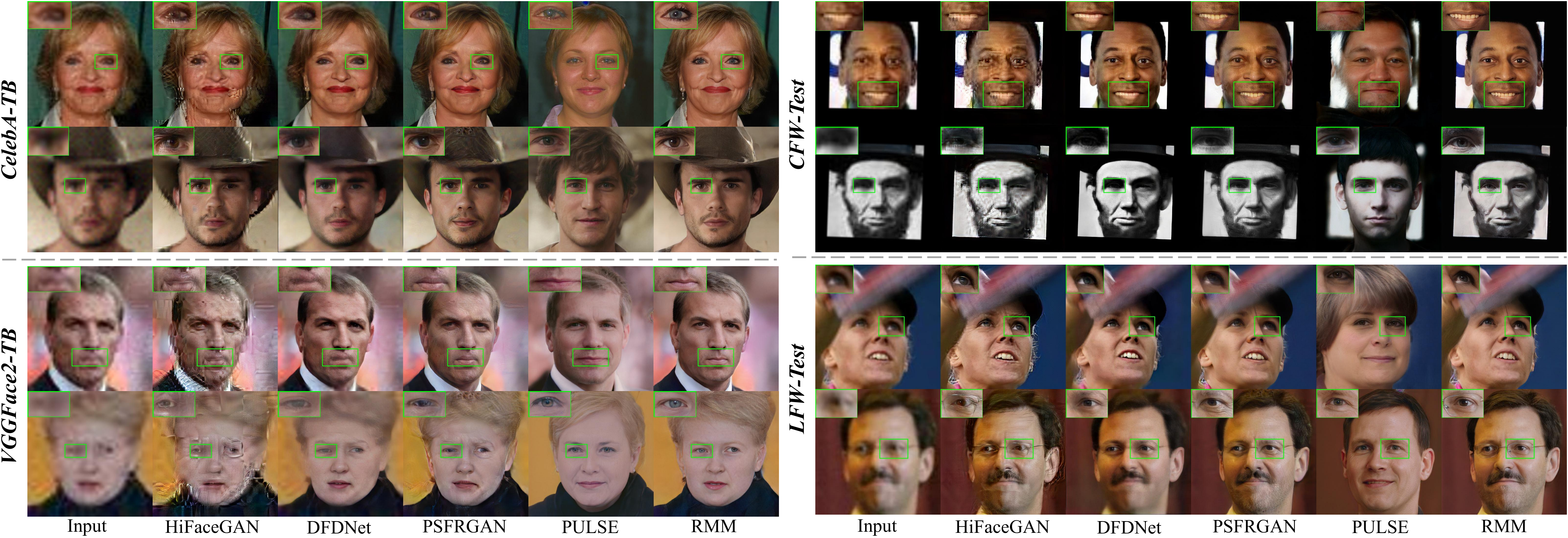}
\end{center}
\vspace{-13.pt}
   \caption{Qualitative comparison with state-of-the-art methods including HiFaceGAN \cite{hifacegan}, DFDNet \cite{dfdnet2020}, PSFRGAN \cite{psfr} and PULSE \cite{pulse2020} on CelebA-TB, VGGFace2-TB, CFW-Test and LFW-Test. Our BFR results have higher fidelity details such as the eyes and mouth. \textbf{Zoom in for the best view}. More results can be found in the supplementary material.}
   \vspace{-13.pt}
\label{fig:com_blind}
\end{figure*}
\begin{figure*}[ht]
\begin{center}
   \includegraphics[width=13cm]{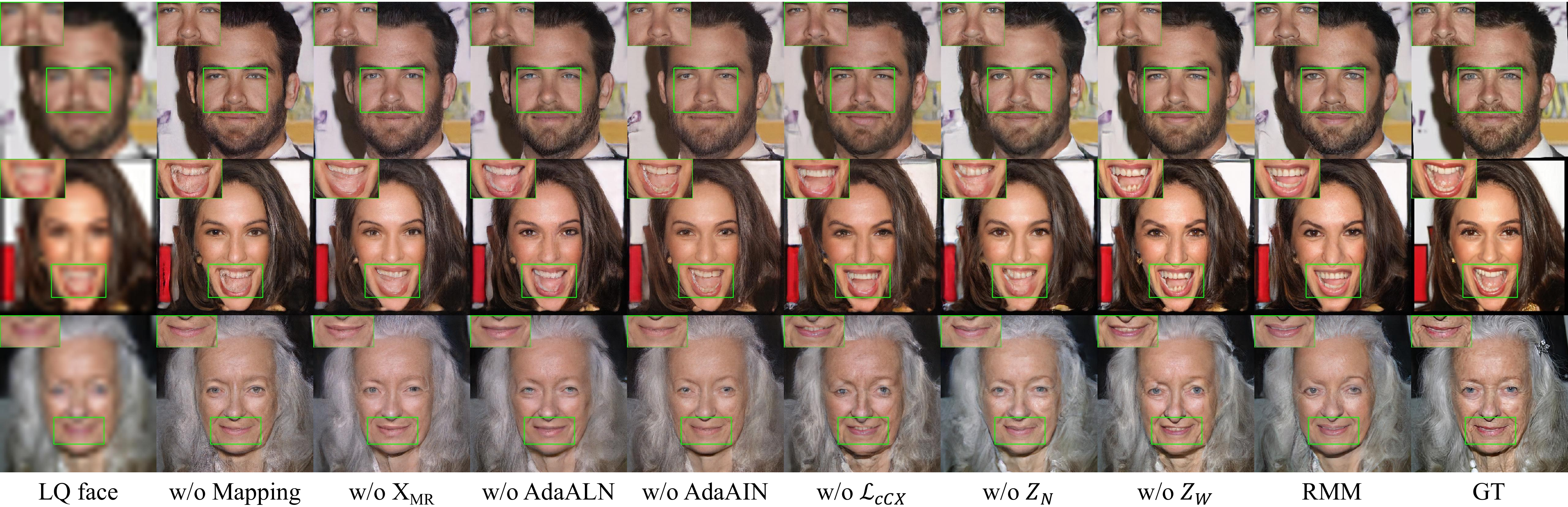}
\end{center}
\vspace{-13.pt}
   \caption{Ablation study of the proposed RMM.}
   \vspace{-11.pt}
\label{fig:ablation}
\end{figure*}
\begin{figure}[h]
\begin{center}
   \includegraphics[width=6cm]{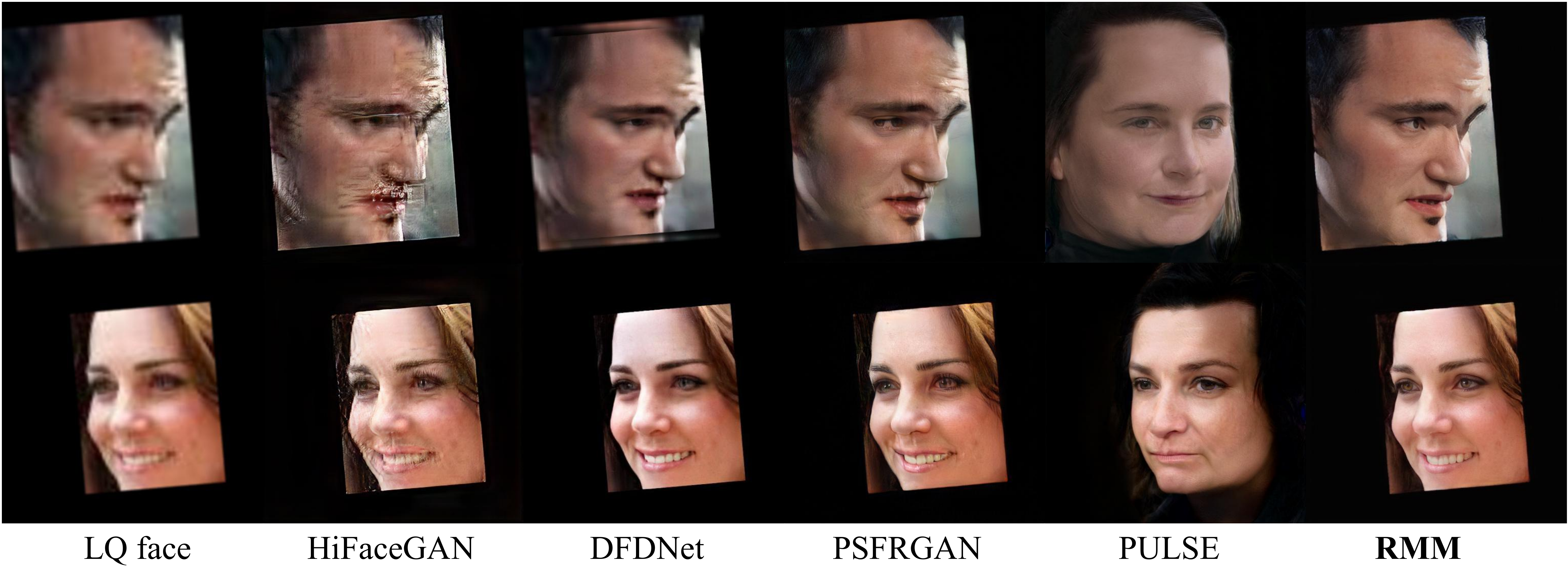}
\end{center}
\vspace{-13.pt}
   \caption{More qualitative results in the large face pose on CFW-Test.}
   \vspace{-13.pt}
\label{fig:large}
\end{figure}

\section{Experiment}
\label{others}
\subsection{Experimental Protocol}
\paragraph{Training Protocol} We train our RMM on the FFHQ \cite{style2019} dataset with 512 size.  The training batch size is set to 8. Each facial component \{\emph{left\_eye, right\_eye, mouth}\} is cropped based on the face landmarks \cite{align}. We train our model with Adam optimizer \cite{adam2014}, and set $\beta_{1}=0.5$, $\beta_{2}=0.999$, learning rate $lr=2e^{-4}$. We employ $D_{i}$ with $i=\{1,2,4,8\}$, and the corresponding weights are 4, 2, 1, 1. We set $\lambda_{rec}=\lambda ^{'}_{rec}=100, \lambda_{cCX}=1$. Moreover, the margin of $\mathcal{L}_{WMM}$ is set to 0.1, the threshold $\eta$ is set to 0.7. The memory size of WMM is set to 982. 

To synthesize the training data that possesses the similar distribution to the real degraded images, we apply the degradation model adopted in \cite{warpnet2018, dfdnet2020, gfp2021},
\begin{equation}
X_{LR}=((X_{HR}\otimes \{\mathbf{k}_{\mathcal{G}}, \mathbf{k}_{\mathcal{M}}\})_{\downarrow r}+\mathbf{n}_{\sigma})_{JPEG_{q}},
\end{equation}

where $\mathbf{k}_{\mathcal{G}}$ denotes Gaussian blur kernel with $\mathcal{G} \in \{1:5\}$, $\mathbf{k}_{\mathcal{M}}$ denotes the motion blur kernels \cite{motion2009}. Moreover, the down-sampling scale r, Gaussian noise $\mathbf{n}_{\sigma}$, and JPEG compression quality $q$ are in the range of $\{2 : 12\}$, $\{1 : 15\}$, and $\{40 : 80\}$, respectively. We randomly integrate these degradation operations in the training stage.

\paragraph{Test Datasets} We test the performance of our RMM with other state-of-the-art methods based on four synthetic datasets and two different real-world datasets as follows.
\begin{itemize}
\item \textbf{CelebA-Test-Blind} is the synthetic dataset with 2,000 CelebA-HQ images from its test set \cite{celeb2017}. The generation way follows the degradation protocol in our training stage. This item is denoted as CelebA-TB.
\item \textbf{CelebA-Test-Downsampling} is the original  2,000 CelebA-HQ images for the test, which is obtained directly by downsampling 4, 8, or 16 times. This item is denoted as CelebA-TD.
\item \textbf{VGGFace2-Test-Blind} is the synthetic dataset with 3,000 test images of VggFace2 \cite{vggface2_2018}. Similarly, it is denoted as VGGFace2-TB.
\item \textbf{VGGFace2-Test-Downsampling} is denoted as VGGFace2-TD. 
\item \textbf{CFW-Test.} IIIT-CFW \cite{cfw} is used in the cartoon face classification and Photo2Cartoon tasks in the wild. We select 252 challenging LQ images with resolution lower than $80 \times 80$.
\item \textbf{LFW-Test.} \cite{lfw} contains LQ images in the wild. We collect 1711 testing images with distinct identities in the validation partition, where the image with the first index is selected.
\end{itemize}

\begin{table*}[t]
\caption{Quantitative evaluation on CelebA-TB, VGGFace2-TB, LFW-Test and CFW-Test. The red and blue values mean the top-1 and top-2 scores. Note that input and GT are not involved in scoring.}
\vspace{-10.pt}
\scalebox{0.38}{
\begin{tabular}{|c|ccccccccc|ccccccccc|ccc|ccc|}

\hline 
Dataset & \multicolumn{9}{c|}{CelebA-TB} & \multicolumn{9}{c|}{VGGFace2-TB} & \multicolumn{3}{c|}{LFW-Test} & \multicolumn{3}{c|}{CFW-Test}\tabularnewline
\hline 
\makecell[c]{Methods} & \makecell[c]{LPIPS$\downarrow$} & \makecell[c]{FID$\downarrow$\\(FFHQ/CelebA)} & \makecell[c]{KID$\times 100 \downarrow$\\(FFHQ/CelebA)} & \makecell[c]{NIQE$\downarrow$} & \makecell[c]{MS-SSIM$\uparrow$} & \makecell[c]{PSNR$\uparrow$} & \makecell[c]{SSIM$\uparrow$} & \makecell[c]{FED$\downarrow$} & LLE$\downarrow$ & \makecell[c]{LPIPS$\downarrow$} & \makecell[c]{FID$\downarrow$\\(FFHQ/VGGFace2)} & \makecell[c]{KID$\times 100 \downarrow$\\(FFHQ/VGGFace2)} & \makecell[c]{NIQE$\downarrow$} & \makecell[c]{MS-SSIM$\uparrow$} & \makecell[c]{PSNR$\uparrow$} & \makecell[c]{SSIM$\uparrow$} & \makecell[c]{FED$\downarrow$} & LLE$\downarrow$ & \makecell[c]{FID$\downarrow$\\(FFHQ)} & \makecell[c]{KID$\times 100 \downarrow$\\(FFHQ)} & NIQE$\downarrow$ & \makecell[c]{FID$\downarrow$\\(FFHQ)} & \makecell[c]{KID$\times 100 \downarrow$\\(FFHQ)} & NIQE$\downarrow$\tabularnewline
\hline 
Input & 0.46 & 175.13 / 84.57 & 16.16 / 7.79 & 8.04 & 0.90 & 25.01 & 0.71 & 0.35 & 1.63 & 0.31 & 137.36 / 25.51 & 12.94 / 1.49 & 8.17 & 0.91 & 28.92 & 0.78 & 0.30 & 1.64 & 131.43 & 13.67 & 7.74 & 335.56 & 34.99 & 9.66\tabularnewline
HiFaceGAN\cite{hifacegan} & 0.37 & 88.74 / 52.76 & 6.49 / 3.83 & \textbf{\textcolor{red}{3.96}} & 0.89 & 23.59 & 0.64 & 0.36 & 1.74 & 0.47 & 66.53 / 88.31 & 4.44 / 8.21 & \textbf{\textcolor{red}{4.02}} & 0.88 & 24.91 & 0.66 & 0.35 & \textbf{\textcolor{blue}{1.91}} & \textbf{\textcolor{blue}{56.75}} & \textbf{\textcolor{blue}{4.02}} & \textbf{\textcolor{red}{3.95}} & 274.09 & 25.23 & \textbf{\textcolor{blue}{4.49}}\tabularnewline
DFDNet\cite{dfdnet2020} & 0.28 & 112.53 / \textbf{\textcolor{blue}{27.57}} & 9.65 / \textbf{\textcolor{blue}{1.51}} & 5.17 & 0.89 & 23.58 & \textbf{\textcolor{red}{0.70}} & \textbf{\textcolor{blue}{0.34}} & 1.68 & \textbf{\textcolor{red}{0.31}} & 66.09 / \textbf{\textcolor{red}{35.86}} & 4.86 / \textbf{\textcolor{red}{2.43}} & 5.50 & 0.88 & 24.41 & \textbf{\textcolor{red}{0.74}} & \textbf{\textcolor{blue}{0.34}} & 2.06 & 77.96 & 6.57 & 5.26 & 272.65 & 26.98 & 6.25\tabularnewline
PSFRGAN\cite{psfr} & \textbf{\textcolor{blue}{0.28}} & \textbf{\textcolor{blue}{74.36}} / 28.97 & 5.01 / 1.71 & \textbf{\textcolor{blue}{4.05}} & \textbf{\textcolor{blue}{0.90}} & \textbf{\textcolor{blue}{24.41}} & 0.67 & 0.35 & \textbf{\textcolor{blue}{1.65}} & 0.39 & \textbf{\textcolor{blue}{41.66}} / 65.45 & \textbf{\textcolor{blue}{2.03}} / 6.09 & \textbf{\textcolor{blue}{4.13}} & \textbf{\textcolor{blue}{0.90}} & \textbf{\textcolor{blue}{26.09}} & 0.70 & 0.35 & 1.95 & 58.69 & 4.04 & \textbf{\textcolor{blue}{4.31}} & 253.77 & 24.54 & 5.11\tabularnewline
PULSE\cite{pulse2020} & 0.41 & 79.55 / 89.58 & \textbf{\textcolor{red}{4.53 }}/ 7.62 & 4.61 & 0.74 & 20.85 & 0.57 & 0.74 & 91.97 & 0.46 & 66.67 / 109.57 & 3.77 / 10.01 & 4.60 & 0.72 & 22.00 & 0.61 & 0.73 & 90.31 & 76.72 & 4.79 & 4.57 & \textbf{\textcolor{red}{134.11}} & \textbf{\textcolor{red}{7.51}} & \textbf{\textcolor{red}{4.47}}\tabularnewline
GLEAN* \cite{bank2021} & 0.35&	99.33/56.90&	9.04/3.04&	5.31&	0.88&	23.16&	0.68&	0.42&	1.90&	\textbf{\textcolor{blue}{0.37}}&	47.22/75.08&	2.52/6.52&	5.40&	0.90&	\textbf{\textcolor{red}{26.38}}&	0.74&	0.38&	1.94&	64.99&	4.13&	5.23&	262.29&	24.17&	5.30\tabularnewline
RMM & \textbf{\textcolor{red}{0.25}} & \textbf{\textcolor{red}{71.88 / 23.49}} & \textbf{\textcolor{blue}{4.85}} / \textbf{\textcolor{red}{1.33}} & 4.19 & \textbf{\textcolor{red}{0.91}} & \textbf{\textcolor{red}{24.80}} & \textbf{\textcolor{blue}{0.69}} & \textbf{\textcolor{red}{0.31}} & \textbf{\textcolor{red}{1.52}} & 0.38 & \textbf{\textcolor{red}{36.47}} / \textbf{\textcolor{blue}{63.52}} & \textbf{\textcolor{red}{1.72}} / \textbf{\textcolor{blue}{5.84}} & 4.33 & \textbf{\textcolor{red}{0.91}} & 26.01 & \textbf{\textcolor{blue}{0.71}} & \textbf{\textcolor{red}{0.32}} & \textbf{\textcolor{red}{1.86}} & \textbf{\textcolor{red}{54.21}} & \textbf{\textcolor{red}{3.65}} & 4.33 & \textbf{\textcolor{blue}{239.90}} & \textbf{\textcolor{blue}{22.77}} & 5.30\tabularnewline
\hline 
GT & 0 & 87.97 / 0 & 6.99 / 0 & 4.62 & 1 & $\infty$ & 1 & 0 & 0 & 0 & 96.16 / 0 & 8.28 / 0 & 6.98 & 1 & $\infty$ & 1 & 0 & 0 & - & - & - & - & - & -\tabularnewline
\hline 
\end{tabular}
}
\vspace{-16.pt}
\label{tab: two}
\end{table*}

\subsection{Comparisons with State-of-the-art Methods}
For BFR task, we employ several evaluation metrics considering the image fidelity (FID \cite{FID}, KID \cite{KID}, NIQE \cite{NIQE}), pixel-wise perception (PSNR, SSIM, MS-SSIM), patch-wise perception (LPIPS \cite{lpips}) and the identity preservation (FED, LLE). Specifically, KID \cite{KID} is the Kernel Inception Distance using the squared Maximum Mean Discrepancy (MMD) with a polynomial kernel, which is used in \cite{ugatit2020,nicegan}. NIQE \cite{NIQE} is a non-reference blind image quality assessment for grayscale images, MS-SSIM considers multi-scale SSIM between the FSR result and the ground truth. Moreover, we measure the feature em-bedding distance (FED) and landmark localization error (LLE) with a pretrained face recognition model based on dlib toolkit. We evaluate FID and KID by calculating the distribution distances between the BFR results and the real images from FFHQ \cite{style2019} as well as the original test set, i.e., CelebA-HQ \cite{celeb2017} or VGGFace2 \cite{vggface2_2018}, respectively.

We compare our RMM with HiFaceGAN \cite{hifacegan}, DFDNet \cite{dfdnet2020}, PSFRGAN \cite{psfr} and PULSE \cite{pulse2020} on the CelebA-TB, VGGFace2-TB, LFW-Test and CFW-Test, as shown in Table \ref{tab: two}. We conduct the evaluation using the same degraded input image for the compared methods, based on the public metric project \cite{metric}. There are no ground-truth HQ images on LFW-Test and CFW-Test, so we only evaluate the non-reference fidelity metrics. On CFW-Test, PULSE \cite{pulse2020} has the best score, but with a serious identity distortion, as shown in Figure \ref{fig:com_blind}. Our RMM has more competitive performance under comprehensive evaluation metrics. 

\begin{figure}[h]
\begin{center}
\vspace{-8.pt}
   \includegraphics[width=4cm]{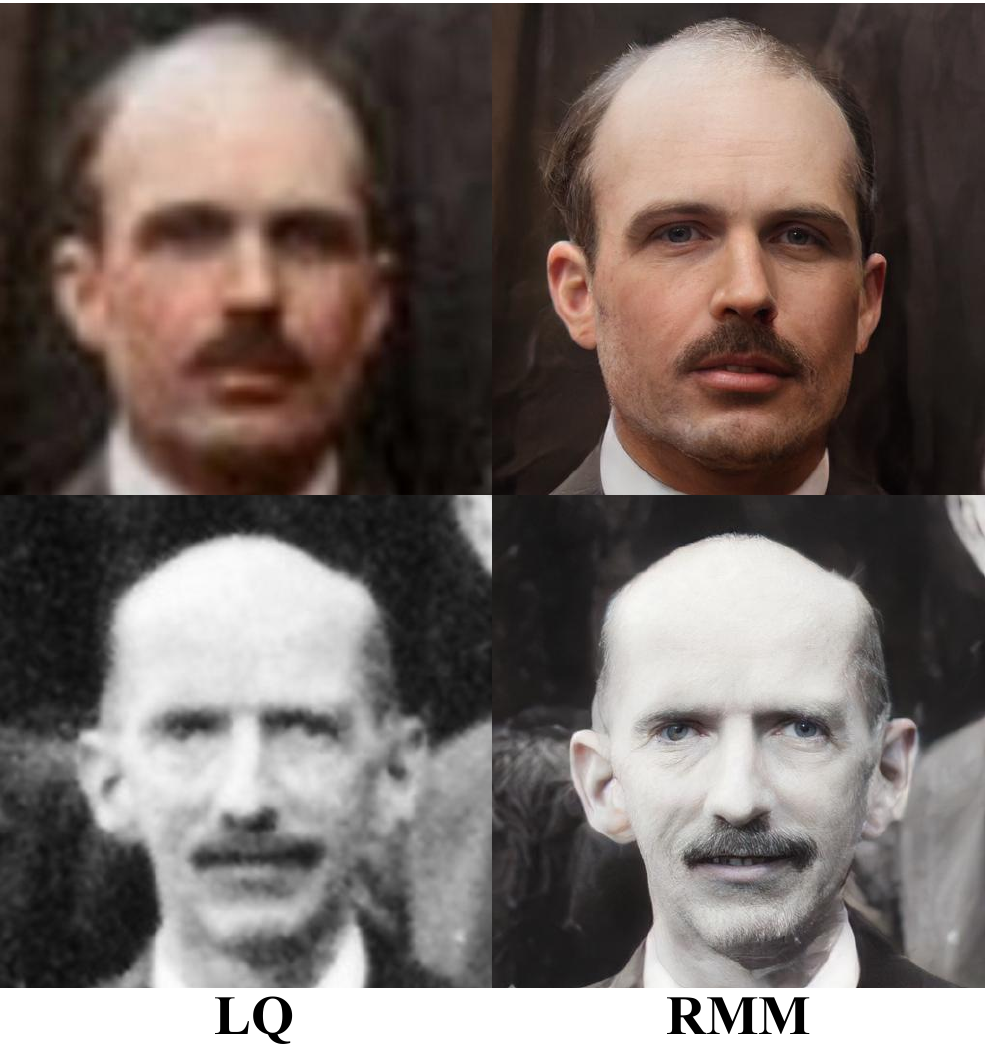}
\end{center}
\vspace{-13.pt}
   \caption{More detailed faces in old photos.}
   \vspace{-13.pt}
\label{fig:big}
\end{figure}

As for qualitative comparison, it is difficult for HiFaceGAN \cite{hifacegan} to handle some challenging degraded images, and there are obvious structural artifacts. The background (e.g., hat, hair) of the results of DFDNet \cite{dfdnet2020} is blurring, as shown in col 3, Figure \ref{fig:com_blind}. Moreover, the eyes and mouth are not high-fidelity, although based on the offline high-quality component dictionary in some challenging cases. PSFRGAN \cite{psfr} has a more stable BFR performance than DFDNet \cite{dfdnet2020}, but with some random artifacts on the image. And the structural and texture details are not captured well in some challenging cases. PULSE \cite{pulse2020} generates BFR faces via the latent space exploration of the pretrained generative models (e.g., StyleGAN \cite{style2019}), the structural information is changed to another unknown identity, as shown in col 5, Figure \ref{fig:com_blind}. Our RMM considers the high-frequency wavelet memory prior and the universal prior against diverse blind degradation patterns. Therefore, the restored faces are higher-quality with structural preservation (e.g., the mouth of row3, the eyeglasses of row 8) and detail restoration (e.g., the eyes and background in col 6). More results are shown in Figure \ref{fig:large} for the large face pose, and Figure \ref{fig:big} for the old photos of the 5-th Solvay conference taken in 1927.
\begin{figure*}[h]
\begin{center}
\vspace{-8.pt}
   \includegraphics[width=15cm]{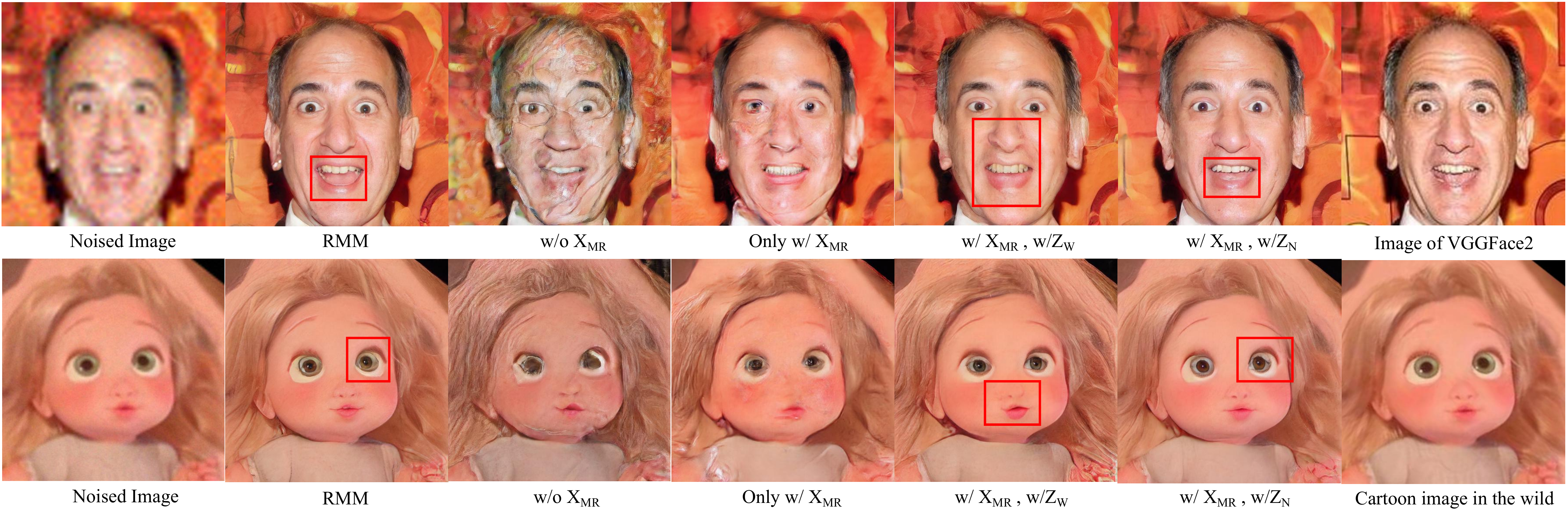}
\end{center}
\vspace{-13.pt}
   \caption{The ablation results in the seriously distortion situations.}
   \vspace{-11.pt}
\label{fig:baby}
\end{figure*}

\subsection{Ablation Study}
The qualitative results are shown in Figure \ref{fig:ablation}. $w/o\ Mapping$ means discarding the mapping network for the input noise $Z$. $w/o\ X_{MR}$ means no refinement of the spatial feature for the input LQ image. $w/o\ AdaALN$ and $w/o\ AdaAIN$ mean only AdaAIN or AdaALN is conducted, respectively. $w/o\ Z_{N}$ denotes that we utilize the spatial feature and wavelet memory to modulate RMM, and  $w/o\ Z_{W}$ means only noise prior and spatial modulations are considered. The results of RMM (col 9) are with higher fidelity than that of $w/o\ Z_{N}$ and $w/o\ Z_{W}$, and can capture more precise texture and structural details, e.g., the shape of eyes, mouth and tooth.

\paragraph{Whether depending on $X_{MR}$} As we all know, $X_{MR}$ is supposed to refine the global spatial information. We conduct more ablation studies, as shown in Figure \ref{fig:baby}. 

(a). If directly using the input degraded spatial feature from $X_{LR}$, denoted as $w/o X_{MR}$, there will be lots of artifacts, especially in the seriously distortion cases. 

(b). If only with $X_{MR}$, i.e., there is no wavelet memory and universal prior modulations, the restored results are messy as well, which demonstrates the essential role of $RM^{3}$ module.

(c). While $w/ X_{MR}$ and $w/ Z_{W}$, it is still challenging for the model to defend the degradation, and some facial components, e.g., eyes or nose, can not be restored correctly.

(d). While $w/ X_{MR}$ and $w/ Z_{N}$, although the corresponding results are reasonable and smooth, they lack some high-frequency details, e.g., the subtle contour of tooth (Figure \ref{fig:ablation}), eyes $\&$ lip (Figure \ref{fig:baby}).

(e). While using the full framework, i.e., RMM, the results are considerable with respect to the global topology and high-frequency details, which is more robust in the wild and heterogeneous domains.
\begin{figure}[ht]
\begin{center}
   \includegraphics[width=7cm]{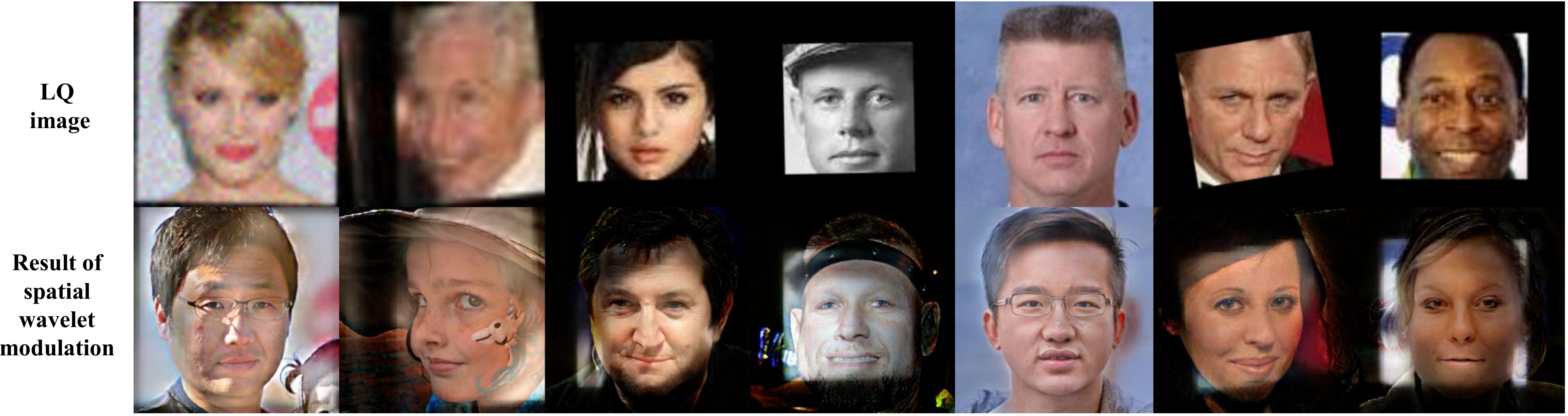}
\end{center}
\vspace{-13.pt}
   \caption{The comparison results with original wavelet feature modulation, which obtained not through average pooling.}
   \vspace{-13.pt}
\label{fig:sp}
\end{figure}
\subsection*{Wavelet Memory Bank} 
\paragraph{Spatial wavelet modulation} The reason of using the latent code not the spatial wavelet feature is to prevent the overfitting. Specifically, the spatial wavelet memory is very helpful to the restoration in the training stage, but causes a bad impact on the samples in the test stage. \textit{No two leaves in the world are exactly the same, and the faces are as well}. If storing the spatial wavelet memory, the matched spatial wavelet memory may have some $wrong$ structural or textural feature, i.e.,  $Z_{S}$ and spatial $Z_{W}$ is inconsistency, which will interfere the topology of the restored image, as shown in Figure \ref{fig:sp}. This will cause a serious identity distortion, e.g., the identity feature embedding distance (FED) is 0.62, and our RMM is 0.31 on CelebA-TB.

\begin{figure}[ht]
\begin{center}
   \includegraphics[width=8.2cm]{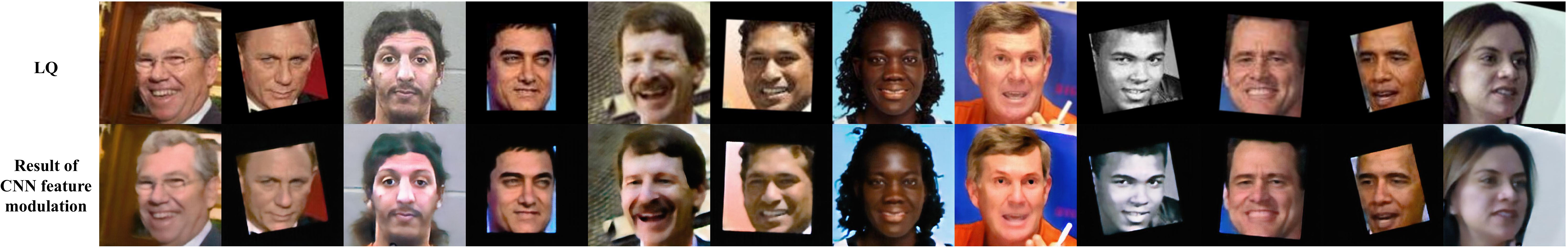}
\end{center}
\vspace{-13.pt}
   \caption{The results of CNN feature modulation based on VGG19.}
   \vspace{-13.pt}
\label{fig:vgg}
\end{figure}

\paragraph{CNN feature modulation} As one of our main contributions, We first propose an effective and efficient Wavelet Memory Module (WMM). The high-frequency wavelet features represent the multi-level local textural details, which helps to restore the texture information that is missing in the degraded image. We have conducted the experiment using deep CNN, i.e., VGG feature as the memory modulation feature, as shown in Figure \ref{fig:vgg}. 

Specifically, the $conv4\_1$ feature is compressed as the memory code by average pooling, and the other modules and loss functions of RMM are maintained. We find the high-frequency information can not be restored at all, even applying the reconstruction loss, the perceptual loss and the multi-scale discriminator. This demonstrates the effectiveness of the proposed wavelet memorized modulation, where the high-frequency wavelet feature is superior beyond the classical deep CNN feature for the high-resolution reconstruction task. 
\vspace{-13.pt}
\paragraph{Memory size}
We conduct the ablation for RMM variants with different memory sizes. Results show that FID, KID and NIQE scores are stable across a wide range of wavelet memory sizes, as shown in Table \ref{tab:mem_size}. $Mem\_100$ has the best FID and KID scores, but with a high NIQE score. Whereas $Mem\_1000$ has a competitive performance of FID and KID, and has the best NIQE score. Furthermore, the larger memory size can contain more wavelet patterns, which is benefit to the model generalization.  
\begin{table}[t]
\centering
\caption{Quantitative evaluation on RMM variants with different memory sizes.}
\resizebox{3.5cm}{1.05cm}{
\begin{tabular}{|c|ccc|ccc|ccc|ccc|ccc|}
\hline 
Dataset & \multicolumn{3}{c|}{Mean}\tabularnewline
\hline 
\makecell[c]{Methods} & \makecell[c]{FID$\downarrow$} & \makecell[c]{KID$\downarrow$} & NIQE$\downarrow$\tabularnewline
\hline 
Mem\_10 & 102.1&	8.8&	4.5343\tabularnewline
Mem\_50 & 102.45&	8.83&	4.5446\tabularnewline
Mem\_100 & \textbf{99.78}&	\textbf{8.42}&	4.7282\tabularnewline
Mem\_200 & 102.61&	8.73&	4.5686\tabularnewline
Mem\_500 & 104.76&	8.84&	4.7031\tabularnewline
Mem\_1000 & 100.38&	8.48&	\textbf{4.5309}\tabularnewline
\hline
\end{tabular}
}
\vspace{-13.pt}
\label{tab:mem_size}
\end{table}

\paragraph{WPD of shallow layer}
We use wavelet packet decomposition of features from shallow layers, i.e., conv1\_1, conv2\_1 and conv3\_1 of VGG19, to modulate RMM. As shown in the bellow table, $RMM_{Conv1\_1}$ has the best NIQE score, but with worse FID and KID, compared with RMM. With WPD of deeper layer, the restored image is more blurred and has bad performance on FID, KID and NIQE scores. On the contrary, RMM uses the WPD of the input image, the structural and textural details are stored to a great extent, to conduct an effective high-frequency feature modulation.
\begin{figure}[h]
\begin{center}
   \includegraphics[width=6cm]{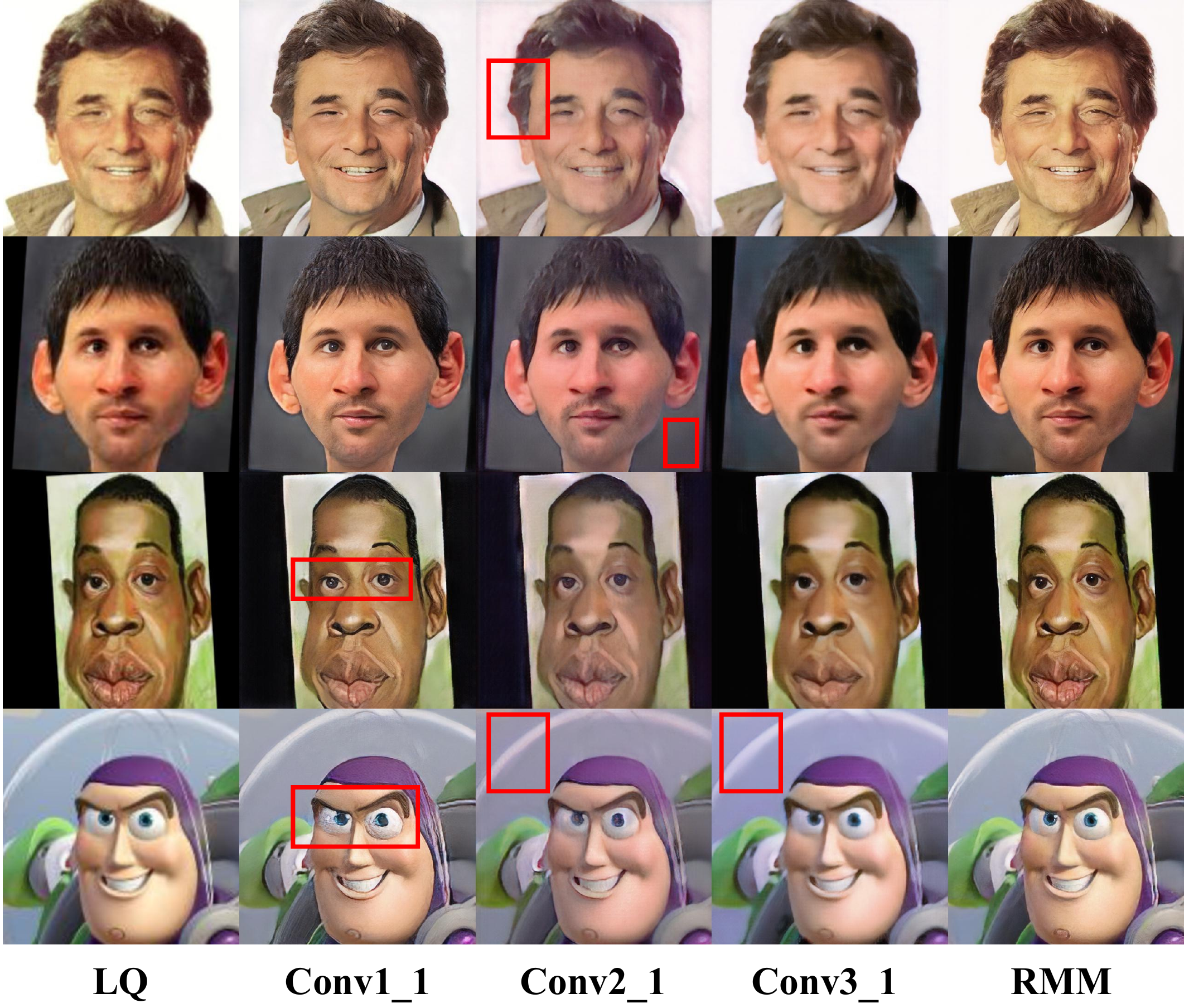}
\end{center}
\vspace{-13.pt}
   \caption{Comparison results of WPD for the shallow layers and the input image. RMM directly uses the WPD of the input face.}
   \vspace{-13.pt}
\label{fig:hi}
\end{figure}
\begin{table}[h]
\centering
\caption{Quantitative evaluation on RMM variants with WPD from different feature layers.}
\resizebox{3.2cm}{0.85cm}{
\begin{tabular}{|c|ccc|}
\hline 
Dataset & \multicolumn{3}{c|}{Mean}\tabularnewline
\hline 
\makecell[c]{Methods} &\makecell[c]{FID$\downarrow$} & \makecell[c]{KID$\downarrow$} & NIQE$\downarrow$\tabularnewline
\hline 
$RMM_{Conv1\_1}$ & 109.41&	9.29&	\textbf{4.29}\tabularnewline
$RMM_{Conv2\_1}$ & 126.04&	11.27&	5.90\tabularnewline
$RMM_{Conv3\_1}$ & 125.54&	11.13&	6.46\tabularnewline
RMM & \textbf{100.61}&	\textbf{8.24}&	4.54\tabularnewline
\hline
\end{tabular}
}

\label{tab:blk}
\vspace{-18.pt}
\end{table}
\vspace{-10.pt}
\begin{figure}[h]
\begin{center}
   \includegraphics[width=7cm]{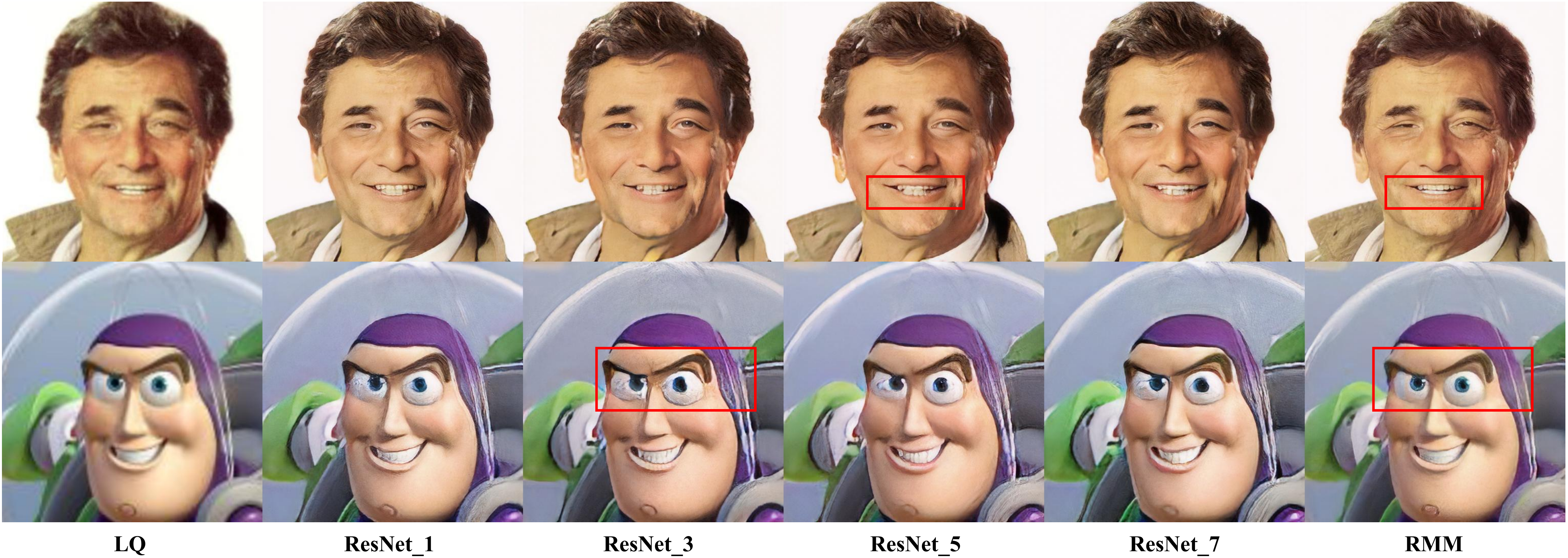}
\end{center}
\vspace{-13.pt}
   \caption{Comparison results for query from different shallow layers.}
   \vspace{-13.pt}
\label{fig:hi}
\end{figure}
\begin{table}[h]
\centering
\caption{Quantitative evaluation on RMM variants with query from different shallow layers.}
\resizebox{3cm}{1cm}{
\begin{tabular}{|c|ccc|}
\hline 
Dataset & \multicolumn{3}{c|}{Mean}\tabularnewline
\hline 
\makecell[c]{Methods} & \makecell[c]{FID$\downarrow$} & \makecell[c]{KID$\downarrow$} & NIQE$\downarrow$\tabularnewline
\hline 
Resnet\_0 & 106.9&	9.11&	4.27\tabularnewline
Resnet\_2 & 106.04&	8.64&	4.38\tabularnewline
Resnet\_4 & 106.1&	8.68&	\textbf{4.24}\tabularnewline
Resnet\_6 & 107.33&	8.72&	4.46\tabularnewline
RMM & \textbf{100.61}&	\textbf{8.24}&	4.54\tabularnewline
\hline
\end{tabular}
}
\vspace{-13.pt}
\label{tab:blk}
\end{table}
\vspace{-13.pt}
\paragraph{Query} The query vector is associated with the low-quality image. We use features from shallow layers as the query, e.g., ResNet18 subsequences with index 0, 2, 4 and 6 are used to produce the query, respectively. Note that our RMM adopts the eighth model sequence followed by adaptive average pooling. These settings have higher FID and KID scores, although with slightly lower NIQE scores. 

As for the visual experiment, we find more artifacts, e.g., wrong and messy details, are synthesized in the eyes, mouth, or background areas. The reason may be that the perception of the query generated from a deep layer, e.g., pool\_5 layer, is more global and comprehensive, so the retrieved high-frequency wavelet features are more accurate and matched with the facial content. However, the query of shallow layers is not capable of representing the spatial feature of LQ image well. Because there are lots of low-level degradations in the LQ image, the spatial query from shallow layers may have more mistakes.
\begin{figure}[h]
\begin{center}
   \includegraphics[width=6cm]{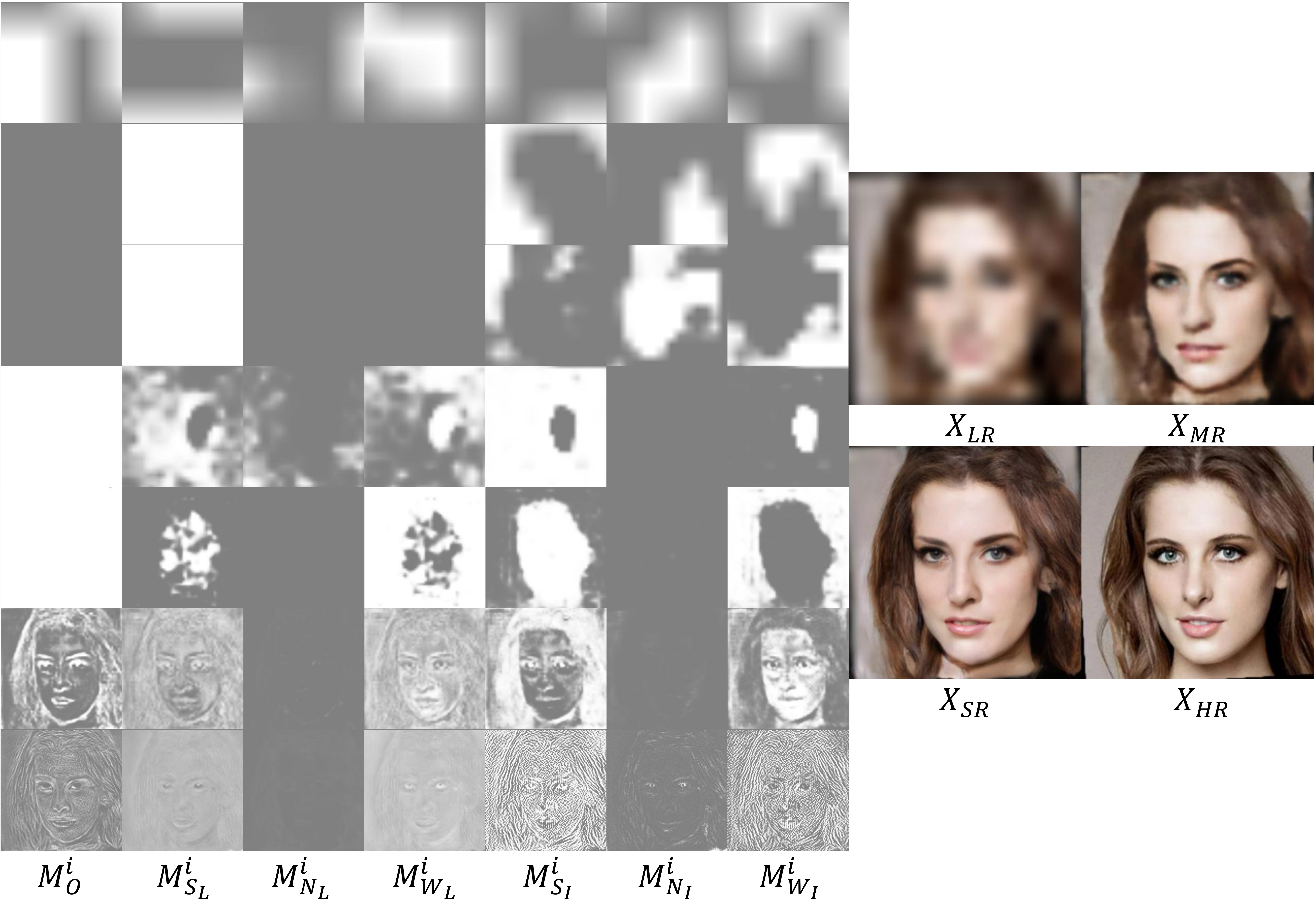}

\end{center}
\vspace{-13.pt}
   \caption{Multi-level attentional maps of RMM.}
   \vspace{-13.pt}
\label{fig:show}
\end{figure}
\subsection{Attentional Map Visualization}
We show the multi-level attentional maps of different decoder features with RMM modulation for the inferred LQ face, in Figure \ref{fig:show}. The bright area means the activation for different feature embeddings. $M_{O}^{i}$ shows that RMM pays more attention on layer denormalization (LDN) in the low-level layers (rows $2\&3\&7$), while the middle layers (row $4\&5$) have more attentions on the instance denormalization (IDN). Face edges (row 6) are also activated in the IDN. The proposed 7 kinds of attentional maps are meaningful for the comprehensive feature fusion. Although there are few activations of $M_{N_{I}}^{i}$ and $M_{N_{L}}^{i}$ for LDN and IDN of the test images, the noise modulation has an essential role for RMM training, as shown in Figure \ref{fig:pipeline} (right) and Figure \ref{fig:ablation} ($w/o\ Z_{W}$, $w/o\ Z_{N}$). 

\subsection{Failure cases}
There are two hard cases. One is for the generalization in heterogeneous domains, if the cartoon face is similar to the real-world face, e.g., the eye size is closed to that of humans, the restored eyes will lose the cartoon style. The other is the face with strange makeup, which makes it so ambiguous that the facial component is wrongly restored.
\begin{figure}[ht]
\begin{center}
   \includegraphics[width=7cm]{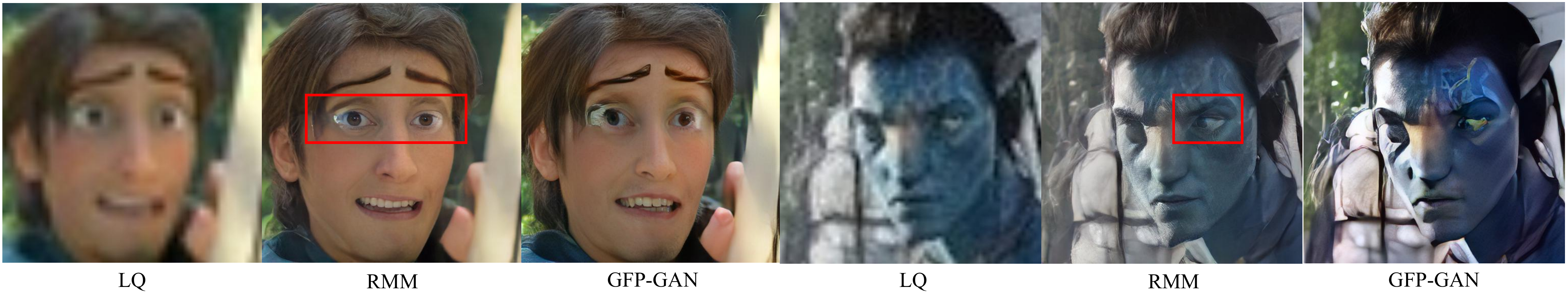}
\end{center}
\vspace{-13.pt}
   \caption{Some failure cases of RMM in the heterogeneous domain. RMM respects the original structural and color distribution of the LQ face.}
   \vspace{-13.pt}
\label{fig:noise} 
\end{figure}

\section{Conclusion}
We propose an RMM framework for blind face restoration. We apply random noise as well as unsupervised wavelet memory to adaptively modulate the BFR generator, considering AdaAO, AdaALN and AdaAIN. RMM controls three embedding modulations respectively associated with the global spatial content, high-frequency texture and structural details, as well as a learnable universal prior against diverse blind image degradation patterns. Experimental results show the superiority of the memorized modulation method on blind face enhancement and a good generalization in the wild and heterogeneous domains, e.g., oil painting, 3D cartoons, pencil drawing, exaggerated drawing and NIR image.

{\small
\bibliographystyle{ieee_fullname}
\bibliography{egbib}

\begin{thebibliography}{10}\itemsep=-1pt

\bibitem{ln2016}
Jimmy~Lei Ba, Jamie~Ryan Kiros, and Geoffrey~E Hinton.
\newblock Layer normalization.
\newblock {\em arXiv preprint arXiv:1607.06450}, 2016.

\bibitem{KID}
Miko{\l}aj Bi{\'n}kowski, Dougal~J Sutherland, Michael Arbel, and Arthur
  Gretton.
\newblock Demystifying mmd gans.
\newblock In {\em ICLR}, 2018.

\bibitem{align}
Adrian Bulat and Georgios Tzimiropoulos.
\newblock How far are we from solving the 2d \& 3d face alignment problem? (and
  a dataset of 230,000 3d facial landmarks).
\newblock In {\em ICCV}, 2017.

\bibitem{vggface2_2018}
Qiong Cao, Li Shen, Weidi Xie, Omkar~M Parkhi, and Andrew Zisserman.
\newblock Vggface2: A dataset for recognising faces across pose and age.
\newblock In {\em 2018 13th IEEE international conference on automatic face \&
  gesture recognition (FG 2018)}, pages 67--74. IEEE, 2018.

\bibitem{bank2021}
Kelvin~CK Chan, Xintao Wang, Xiangyu Xu, Jinwei Gu, and Chen~Change Loy.
\newblock Glean: Generative latent bank for large-factor image
  super-resolution.
\newblock {\em arXiv preprint arXiv:2012.00739}, 2020.

\bibitem{psfr}
Chaofeng Chen, Xiaoming Li, Lingbo Yang, Xianhui Lin, Lei Zhang, and Kwan-Yee~K
  Wong.
\newblock Progressive semantic-aware style transformation for blind face
  restoration.
\newblock {\em arXiv preprint arXiv:2009.08709}, 2020.

\bibitem{nicegan}
Runfa Chen, Wenbing Huang, Binghui Huang, Fuchun Sun, and Bin Fang.
\newblock Reusing discriminators for encoding: Towards unsupervised
  image-to-image translation.
\newblock In {\em CVPR}, pages 8168--8177, 2020.

\bibitem{m1}
Cesc Chunseong~Park, Byeongchang Kim, and Gunhee Kim.
\newblock Attend to you: Personalized image captioning with context sequence
  memory networks.
\newblock In {\em CVPR}, pages 895--903, 2017.

\bibitem{pool2020}
Shuguang Cui.
\newblock Towards content-independent multi-reference super-resolution:
  Adaptive pattern matching and feature aggregation.
\newblock 2020.

\bibitem{example2019}
Berk Dogan, Shuhang Gu, and Radu Timofte.
\newblock Exemplar guided face image super-resolution without facial landmarks.
\newblock In {\em CVPR Workshops}, pages 0--0, 2019.

\bibitem{com1}
Chao Dong, Yubin Deng, Chen~Change Loy, and Xiaoou Tang.
\newblock Compression artifacts reduction by a deep convolutional network.
\newblock In {\em ICCV}, pages 576--584, 2015.

\bibitem{sr1}
Chao Dong, Chen~Change Loy, Kaiming He, and Xiaoou Tang.
\newblock Learning a deep convolutional network for image super-resolution.
\newblock In {\em ECCV}, pages 184--199. Springer, 2014.

\bibitem{denoise2}
Majed El~Helou, Ruofan Zhou, and Sabine S{\"u}sstrunk.
\newblock Stochastic frequency masking to improve super-resolution and
  denoising networks.
\newblock In {\em ECCV}, pages 749--766. Springer, 2020.

\bibitem{com2}
Jun Guo and Hongyang Chao.
\newblock Building dual-domain representations for compression artifacts
  reduction.
\newblock In {\em ECCV}, pages 628--644. Springer, 2016.

\bibitem{resnet}
Kaiming He, Xiangyu Zhang, Shaoqing Ren, and Jian Sun.
\newblock Deep residual learning for image recognition.
\newblock In {\em CVPR}, pages 770--778, 2016.

\bibitem{FID}
M. Heusel, H. Ramsauer, T. Unterthiner, B. Nessler, and S. Hochreiter.
\newblock Gans trained by a two time-scale update rule converge to a local nash
  equilibrium.
\newblock In {\em NIPS}, 2017.

\bibitem{lfw}
Gary~B Huang, Marwan Mattar, Tamara Berg, and Eric Learned-Miller.
\newblock Labeled faces in the wild: A database forstudying face recognition in
  unconstrained environments.
\newblock In {\em Workshop on faces in'Real-Life'Images: detection, alignment,
  and recognition}, 2008.

\bibitem{wavelet2017}
Huaibo Huang, Ran He, Zhenan Sun, and Tieniu Tan.
\newblock Wavelet-srnet: A wavelet-based cnn for multi-scale face super
  resolution.
\newblock In {\em ICCV}, pages 1689--1697, 2017.

\bibitem{metric_color}
Janspiry.
\newblock Color similarity calculation.
\newblock \url{https://github.com/Janspiry/Color-Similarity-Calculation}.

\bibitem{celeb2017}
Tero Karras, Timo Aila, Samuli Laine, and Jaakko Lehtinen.
\newblock Progressive growing of gans for improved quality, stability, and
  variation.
\newblock {\em arXiv preprint arXiv:1710.10196}, 2017.

\bibitem{style2019}
Tero Karras, Samuli Laine, and Timo Aila.
\newblock A style-based generator architecture for generative adversarial
  networks.
\newblock In {\em CVPR}, pages 4401--4410, 2019.

\bibitem{stylegan2_2020}
Tero Karras, Samuli Laine, Miika Aittala, Janne Hellsten, Jaakko Lehtinen, and
  Timo Aila.
\newblock Analyzing and improving the image quality of stylegan.
\newblock In {\em CVPR}, pages 8110--8119, 2020.

\bibitem{ugatit2020}
Junho Kim, Minjae Kim, Hyeonwoo Kang, and Kwanghee Lee.
\newblock U-gat-it: unsupervised generative attentional networks with adaptive
  layer-instance normalization for image-to-image translation.
\newblock In {\em ICLR}, 2020.

\bibitem{adam2014}
Diederik~P Kingma and Jimmy Ba.
\newblock Adam: A method for stochastic optimization.
\newblock {\em arXiv preprint arXiv:1412.6980}, 2014.

\bibitem{deblur2}
Orest Kupyn, Volodymyr Budzan, Mykola Mykhailych, Dmytro Mishkin, and
  Ji{\v{r}}{\'\i} Matas.
\newblock Deblurgan: Blind motion deblurring using conditional adversarial
  networks.
\newblock In {\em CVPR}, pages 8183--8192, 2018.

\bibitem{motion2009}
Anat Levin, Yair Weiss, Fredo Durand, and William~T Freeman.
\newblock Understanding and evaluating blind deconvolution algorithms.
\newblock In {\em CVPR}, pages 1964--1971. IEEE, 2009.

\bibitem{anigan2021}
Bing Li, Yuanlue Zhu, Yitong Wang, Chia-Wen Lin, Bernard Ghanem, and Linlin
  Shen.
\newblock Anigan: Style-guided generative adversarial networks for unsupervised
  anime face generation.
\newblock {\em arXiv preprint arXiv:2102.12593}, 2021.

\bibitem{faceshifter_2020}
Lingzhi Li, Jianmin Bao, Hao Yang, Dong Chen, and Fang Wen.
\newblock Advancing high fidelity identity swapping for forgery detection.
\newblock In {\em CVPR}, June 2020.

\bibitem{dfdnet2020}
Xiaoming Li, Chaofeng Chen, Shangchen Zhou, Xianhui Lin, Wangmeng Zuo, and Lei
  Zhang.
\newblock Blind face restoration via deep multi-scale component dictionaries.
\newblock In {\em ECCV}, pages 399--415. Springer, 2020.

\bibitem{exemplar2020}
Xiaoming Li, Wenyu Li, Dongwei Ren, Hongzhi Zhang, Meng Wang, and Wangmeng Zuo.
\newblock Enhanced blind face restoration with multi-exemplar images and
  adaptive spatial feature fusion.
\newblock In {\em CVPR}, pages 2706--2715, 2020.

\bibitem{warpnet2018}
Xiaoming Li, Ming Liu, Yuting Ye, Wangmeng Zuo, Liang Lin, and Ruigang Yang.
\newblock Learning warped guidance for blind face restoration.
\newblock In {\em ECCV}, pages 272--289, 2018.

\bibitem{sr2}
Ding Liu, Bihan Wen, Yuchen Fan, Chen~Change Loy, and Thomas~S Huang.
\newblock Non-local recurrent network for image restoration.
\newblock In {\em NIPS}, volume~31. Curran Associates, Inc., 2018.

\bibitem{wave2018}
Pengju Liu, Hongzhi Zhang, Kai Zhang, Liang Lin, and Wangmeng Zuo.
\newblock Multi-level wavelet-cnn for image restoration.
\newblock In {\em CVPR workshops}, pages 773--782, 2018.

\bibitem{masa2021}
Xin Tao Jiangbo Lu Jiaya~Jia Liying~Lu1, Wenbo~Li1.
\newblock Masa-sr: Matching acceleration and spatial adaptation for
  reference-based image super-resolution.
\newblock In {\em CVPR}, 2021.

\bibitem{metric}
Lotayou.
\newblock Metrics\_package.
\newblock \url{https://github.com/Lotayou/Face-Renovation}.

\bibitem{cx2018}
Roey Mechrez, Itamar Talmi, and Lihi Zelnik-Manor.
\newblock The contextual loss for image transformation with non-aligned data.
\newblock In {\em ECCV}, pages 768--783, 2018.

\bibitem{pulse2020}
Sachit Menon, Alexandru Damian, Shijia Hu, Nikhil Ravi, and Cynthia Rudin.
\newblock Pulse: Self-supervised photo upsampling via latent space exploration
  of generative models.
\newblock In {\em CVPR}, pages 2437--2445, 2020.

\bibitem{m3}
Jiaxu Miao, Yunchao Wei, and Yi Yang.
\newblock Memory aggregation networks for efficient interactive video object
  segmentation.
\newblock In {\em CVPR}, pages 10366--10375, 2020.

\bibitem{cfw}
Ashutosh Mishra, Shyam~Nandan Rai, Anand Mishra, and CV Jawahar.
\newblock Iiit-cfw: A benchmark database of cartoon faces in the wild.
\newblock In {\em ECCV}, pages 35--47. Springer, 2016.

\bibitem{NIQE}
Anish Mittal, Rajiv Soundararajan, and Alan~C Bovik.
\newblock Making a “completely blind” image quality analyzer.
\newblock {\em IEEE Signal processing letters}, 20(3):209--212, 2012.

\bibitem{spade2019}
Taesung Park, Ming-Yu Liu, Ting-Chun Wang, and Jun-Yan Zhu.
\newblock Semantic image synthesis with spatially-adaptive normalization.
\newblock In {\em CVPR}, pages 2337--2346, 2019.

\bibitem{unet2015}
Olaf Ronneberger, Philipp Fischer, and Thomas Brox.
\newblock U-net: Convolutional networks for biomedical image segmentation.
\newblock In {\em International Conference on Medical image computing and
  computer-assisted intervention}, pages 234--241. Springer, 2015.

\bibitem{vgg2015}
Karen Simonyan and Andrew Zisserman.
\newblock Very deep convolutional networks for large-scale image recognition.
\newblock In {\em ICLR}, 2015.

\bibitem{in2016}
Dmitry Ulyanov, Andrea Vedaldi, and Victor Lempitsky.
\newblock Instance normalization: The missing ingredient for fast stylization.
\newblock {\em arXiv preprint arXiv:1607.08022}, 2016.

\bibitem{p2phd}
Ting-Chun Wang, Ming-Yu Liu, Jun-Yan Zhu, Andrew Tao, Jan Kautz, and Bryan
  Catanzaro.
\newblock High-resolution image synthesis and semantic manipulation with
  conditional gans.
\newblock In {\em CVPR}, pages 8798--8807, 2018.

\bibitem{gfp2021}
Xintao Wang, Yu Li, Honglun Zhang, and Ying Shan.
\newblock Towards real-world blind face restoration with generative facial
  prior.
\newblock {\em arXiv preprint arXiv:2101.04061}, 2021.

\bibitem{gfpgan}
Xintao Wang, Yu Li, Honglun Zhang, and Ying Shan.
\newblock Towards real-world blind face restoration with generative facial
  prior.
\newblock In {\em The IEEE Conference on Computer Vision and Pattern
  Recognition (CVPR)}, 2021.

\bibitem{memory2015}
Jason Weston, Sumit Chopra, and Antoine Bordes.
\newblock Memory networks.
\newblock In {\em ICLR}, 2015.

\bibitem{invert}
Mingqing Xiao, Shuxin Zheng, Chang Liu, Yaolong Wang, Di He, Guolin Ke, Jiang
  Bian, Zhouchen Lin, and Tie-Yan Liu.
\newblock Invertible image rescaling.
\newblock In {\em ECCV}, pages 126--144. Springer, 2020.

\bibitem{deblur1}
Li Xu, Jimmy~S Ren, Ce Liu, and Jiaya Jia.
\newblock Deep convolutional neural network for image deconvolution.
\newblock {\em NIPS}, 27:1790--1798, 2014.

\bibitem{hifacegan}
Lingbo Yang, Shanshe Wang, Siwei Ma, Wen Gao, Chang Liu, Pan Wang, and Peiran
  Ren.
\newblock Hifacegan: Face renovation via collaborative suppression and
  replenishment.
\newblock In {\em ACM MM}, pages 1551--1560, 2020.

\bibitem{denoise1}
Kai Zhang, Wangmeng Zuo, Yunjin Chen, Deyu Meng, and Lei Zhang.
\newblock Beyond a gaussian denoiser: Residual learning of deep cnn for image
  denoising.
\newblock {\em IEEE transactions on image processing}, 26(7):3142--3155, 2017.

\bibitem{lpips}
Richard Zhang, Phillip Isola, Alexei~A Efros, Eli Shechtman, and Oliver Wang.
\newblock The unreasonable effectiveness of deep features as a perceptual
  metric.
\newblock In {\em CVPR}, pages 586--595, 2018.

\bibitem{sr3}
Yulun Zhang, Kunpeng Li, Kai Li, Lichen Wang, Bineng Zhong, and Yun Fu.
\newblock Image super-resolution using very deep residual channel attention
  networks.
\newblock In {\em ECCV}, pages 286--301, 2018.

\bibitem{copy2020}
Yang Zhang, Ivor~W Tsang, Yawei Luo, Chang-Hui Hu, Xiaobo Lu, and Xin Yu.
\newblock Copy and paste gan: Face hallucination from shaded thumbnails.
\newblock In {\em CVPR}, pages 7355--7364, 2020.

\bibitem{m2}
Minfeng Zhu, Pingbo Pan, Wei Chen, and Yi Yang.
\newblock Dm-gan: Dynamic memory generative adversarial networks for
  text-to-image synthesis.
\newblock In {\em CVPR}, pages 5802--5810, 2019.

\end{thebibliography}
}

\clearpage
\section*{Supplementary material}

\subsection*{Additional Experimental Results} 
\begin{figure*}[ht]
\begin{center}
   \includegraphics[width=15cm]{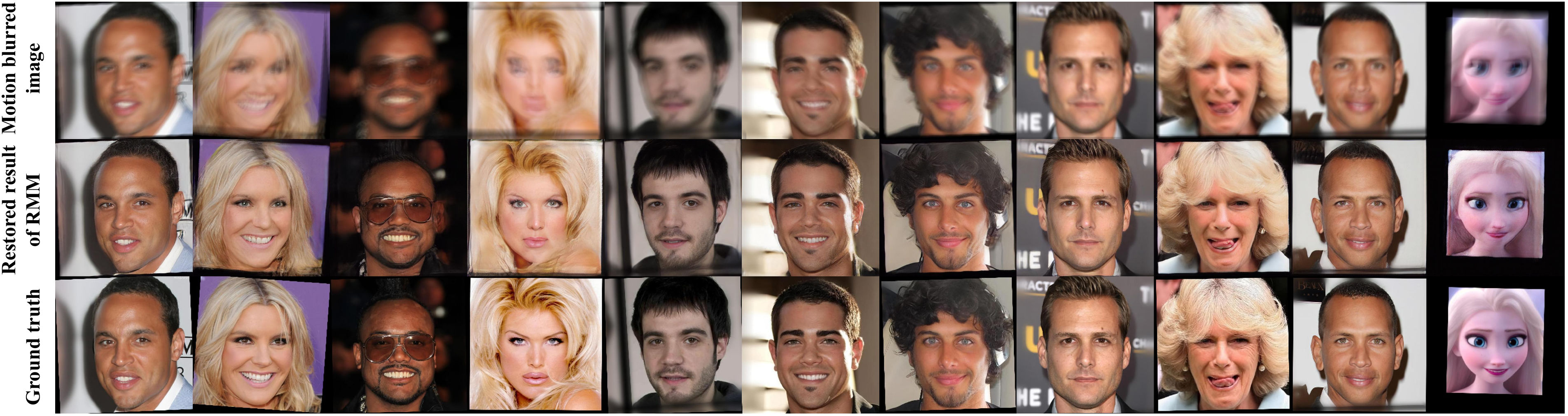}
\end{center}
\vspace{-13.pt}
   \caption{More results on CelebA and the heterogeneous domain (the last) considering the motion blur. \textbf{Zoom in for the best view.}}
   \vspace{-13.pt}
\label{fig:blur}
\end{figure*}

\begin{figure*}[h]
\begin{center}
   \includegraphics[width=15cm]{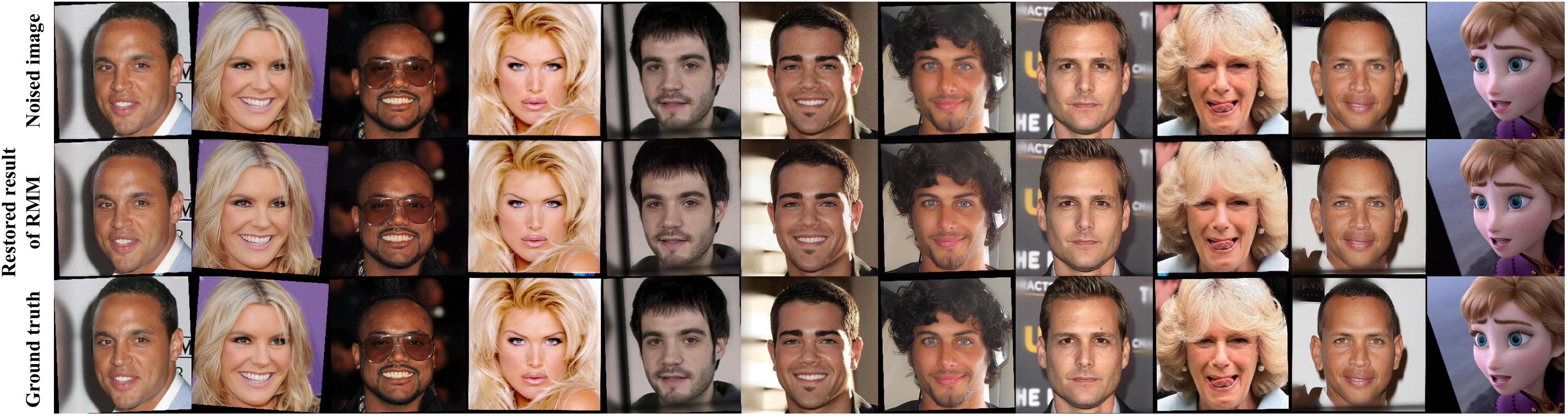}
\end{center}
\vspace{-13.pt}
   \caption{More results on CelebA and the heterogeneous domain (the last) considering the random noise. \textbf{Zoom in for the best view.}}
   \vspace{-13.pt}
\label{fig:noi}
\end{figure*}

\paragraph{Motion blur}
Figure \ref{fig:blur} shows the test images with motion blur, which demonstrates that our RMM can handle casual motion blur degradation, even in the heterogeneous domain. 
\paragraph{Noise degradation factor}
Figure \ref{fig:noi} shows the test images with noise degradation, which demonstrates that our RMM can handle random noise degradation, even in the heterogeneous domain.

\begin{figure}[ht]
\begin{center}
   \includegraphics[width=8cm]{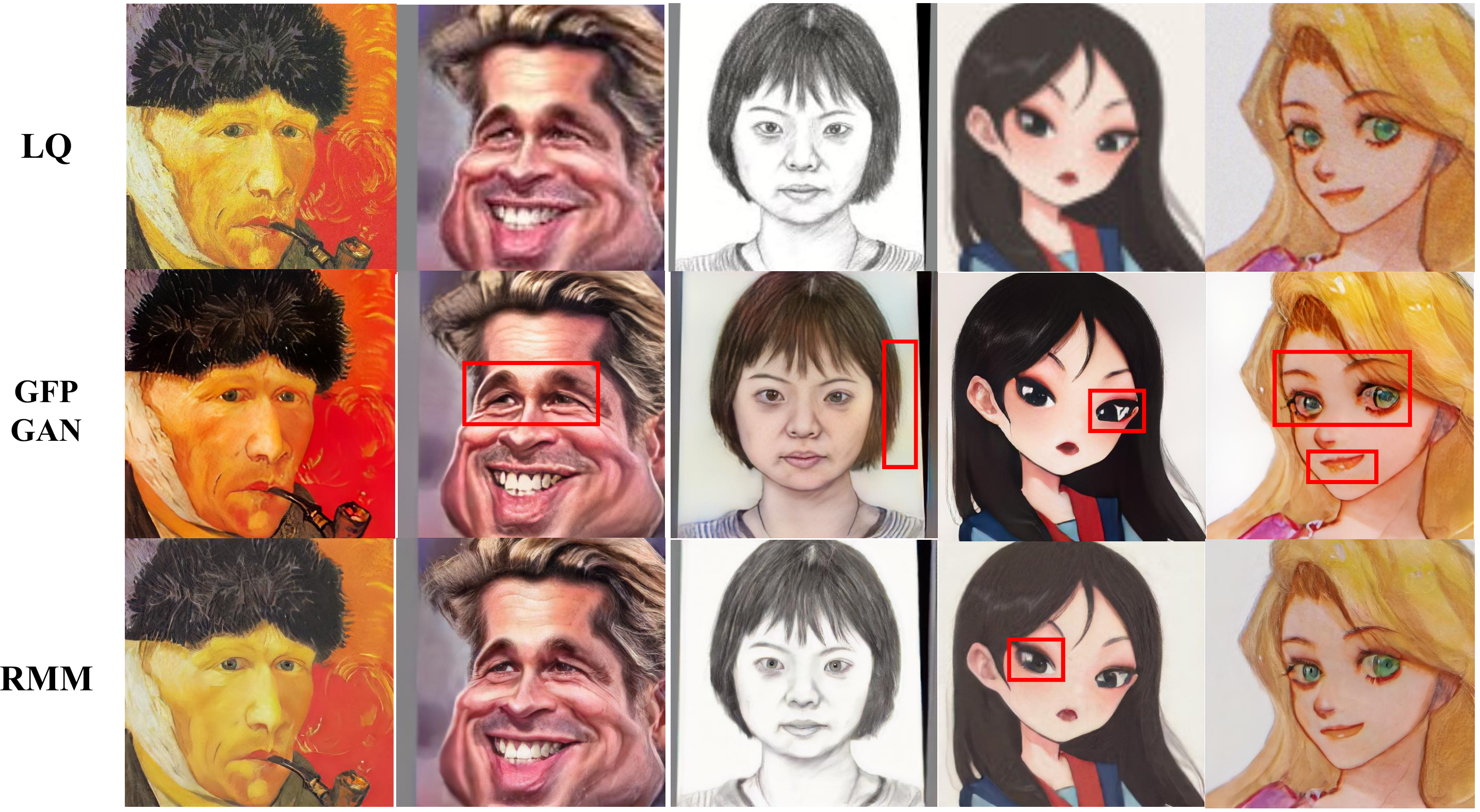}
\end{center}
\vspace{-13.pt}
   \caption{More examples of GFP-GAN and RMM in the wild, including the oil painting, 3D cartoons, pencil drawing and exaggerated drawing. \textbf{Zoom in for the best view.}}
\label{fig:compare2}
\end{figure}

\paragraph{GFP-GAN} We compare with the sota method GFP-GAN. Although our RMM does not depend on the pretrained StyleGAN prior, our performance is also competitive. We show more comparison results for the heterogeneous domains (oil painting, 3D cartoons, pencil drawing and exaggerated drawing) in Figure \ref{fig:compare2}. 

It should be noted that our results preserve the image color well, whereas the restored faces of GFP-GAN have a larger color gap to the original LQ faces, as shown in Table  \ref{tab:color}.
\begin{table}[h]
\centering
\caption{Color similarity scores of GFP-GAN and RMM based on Hist-CIEDE \cite{metric_color}. The smaller score is the best, which indicates the restored face has a higher color fidelity of the LQ image.}
\resizebox{5cm}{0.5cm}{
\begin{tabular}{|c|c|c|c|c|}
\hline 
Dataset & \multicolumn{1}{c|}{CelebA-TB} & \multicolumn{1}{c|}{VGGFace2-TB} & \multicolumn{1}{c|}{LFW-Test} & \multicolumn{1}{c|}{CFW-Test}\tabularnewline
\hline 

GFP-GAN\cite{gfpgan} & 9.97 & 10.34 & 7.51 & 22.72 \tabularnewline
RMM &  \textbf{\textcolor{red}{8.51}} & \textbf{\textcolor{red}{8.30}} & \textbf{\textcolor{red}{6.24
}} & \textbf{\textcolor{red}{17.99}} \tabularnewline
\hline 
\end{tabular}
}
\label{tab:color}
\end{table}
\begin{figure}[ht]
\begin{center}
   \includegraphics[width=8cm]{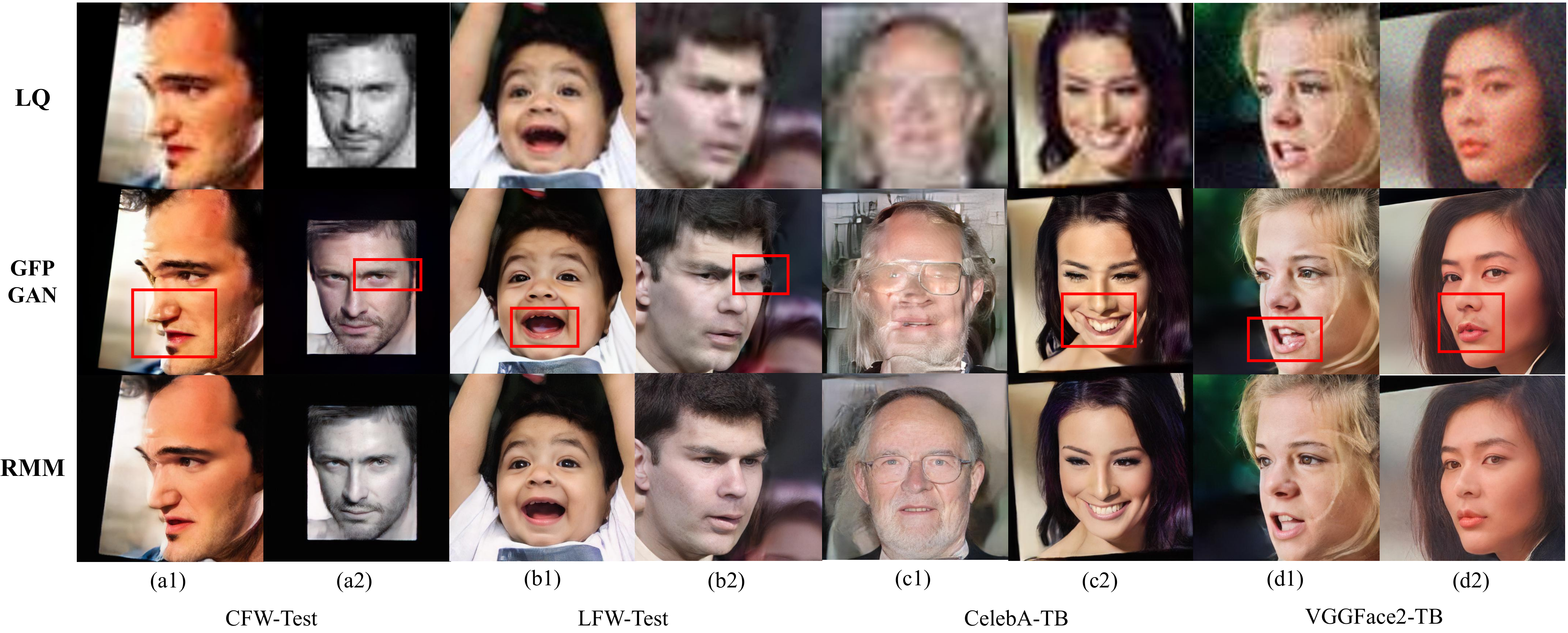}
\end{center}
\vspace{-13.pt}
   \caption{Some restored examples of GFP-GAN and RMM. \textbf{Zoom in for the best view.}}
\label{fig:compare1}
\end{figure}

\begin{figure}[h]
\begin{center}
   \includegraphics[width=8cm]{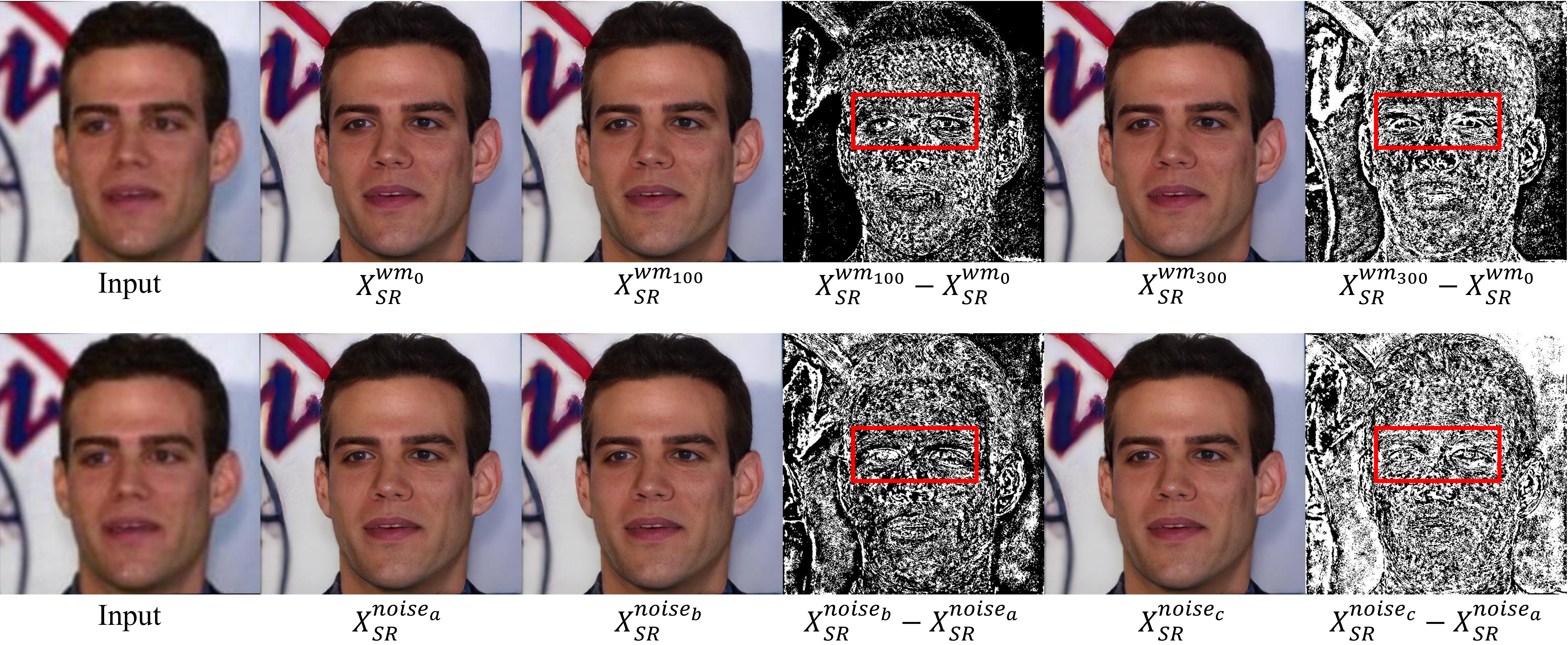}
\end{center}
\vspace{-13.pt}
   \caption{Additional analysis of RMM. The up and below are the results using different wavelet memories, and noises, respectively. Note that $wm_{100}$ means RMM with top-100 wavelet memory, and noise$_{a}$ means one randomly sampled Gaussian noise. \textbf{Zoom in for the best view.}}
\label{fig:ana}
\end{figure}

\begin{figure}[ht]
\begin{center}
   \includegraphics[width=7cm]{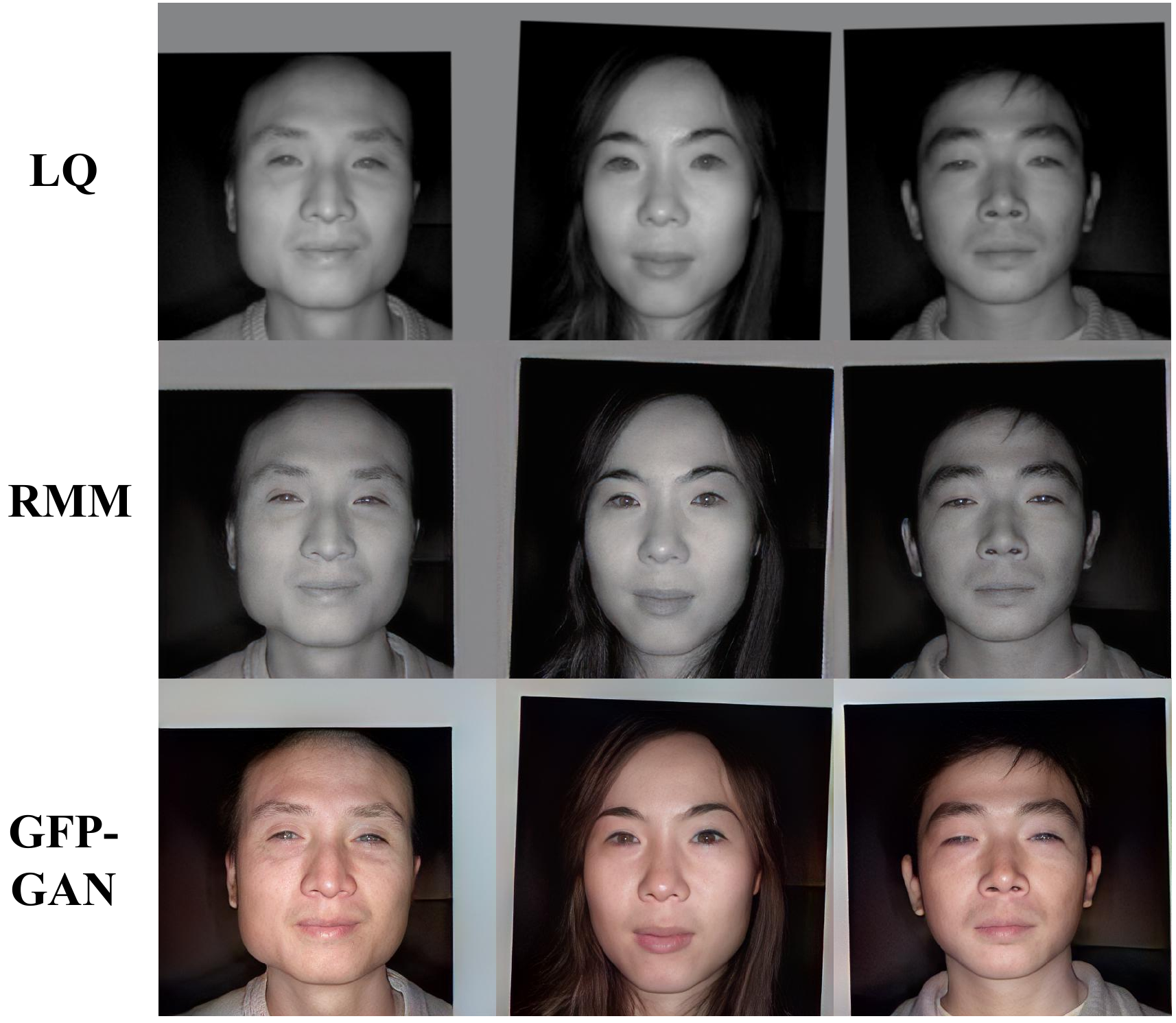}
\end{center}
\vspace{-13.pt}
   \caption{Comparison results in the NIR portrait.}
\label{fig:NIR}
\end{figure}

\paragraph{Color consistency}
Coloring the grayscale image may not be the drawback of GFP-GAN, which is considerable in the photographic domain. However, it is a fact that there is a larger color gap between the restored and original LQ image for GFP-GAN. Especially it lacks controllability in the heterogeneous domains, e.g., the color information of the colored face and the original background will be inconsistent in the sketch or grayscale domain. When it comes to another challenging case, e.g., NIR image enhancement, the lighting information on the face is supposed to be maintained. Otherwise, the scene presentation looks unharmonious, e.g., the NIR portrait seems daytime VIS portrait, as shown in Figure \ref{fig:NIR}.

\begin{figure}[h]
\begin{center}
   \includegraphics[width=8cm]{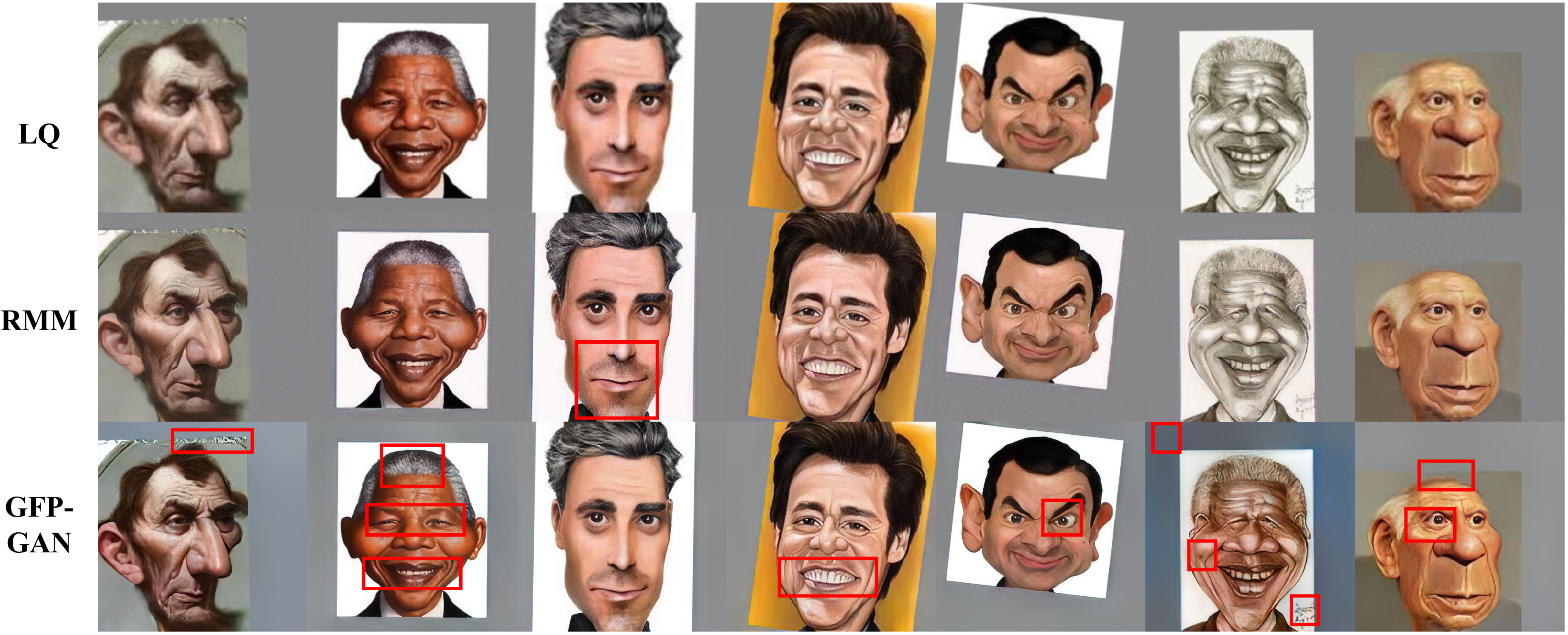} \includegraphics[width=8cm]{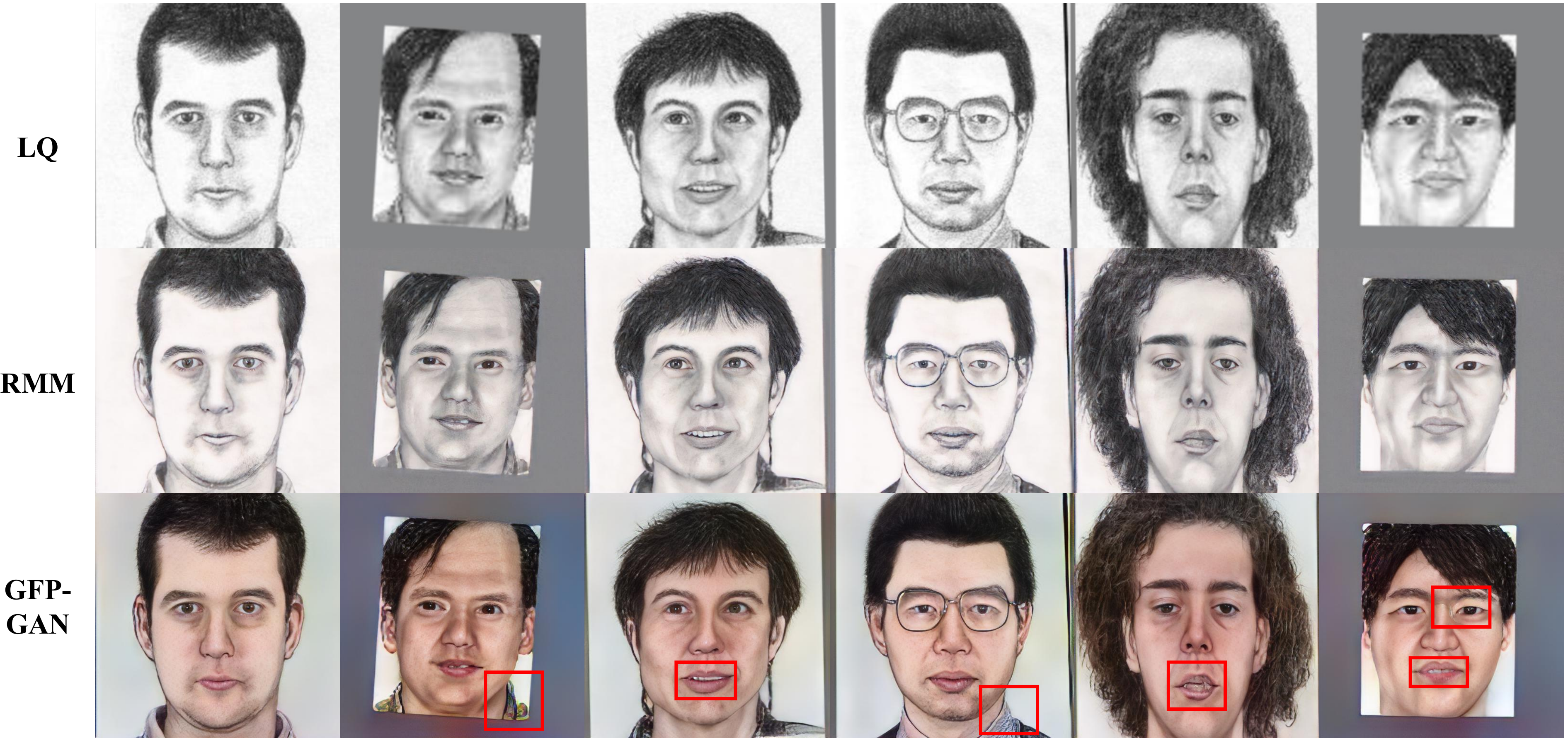}
\end{center}
\vspace{-13.pt}
   \caption{More comparison results in heterogeneous domains, where the upper is in the exaggerated drawing domain, and the bellow is in the sketch domain.}
\label{fig:ex_sk}
\end{figure}

On one hand, the colored restored face may look more pleasing. On the other hand, GFP-GAN does not respect the original color attribute well. Our RMM has a competitive performance compared with GFP-GAN, and better color controllability. Moreover, RMM  has a better generalization in the wild. There are more accurate structural and textural facial details, but fewer artifacts, thanks to the effectiveness of the wavelet memory and the universal prior, as shown in Figure \ref{fig:ex_sk}. 

\paragraph{Generalization in heterogeneous domains}
There are three important aspects for the generalization in heterogeneous domains. The first is the preservation of the global structural and color information from the LQ image, the second is the accurate high-frequency textual restoration using the memorized wavelet modulation, and the last is the ability to defend diverse blind degradations using the learned universal restoration prior. 

Because of these functions, our RMM respects the global attributes of the original image (e.g., in the sketch, grayscale, or oil drawing domain), complements accurate high-frequency local details (e.g., in the 3D cartoon or exaggerated drawing domain), and improves the model robustness and controllability in the wild. RMM realizes reasonable and high-fidelity facial restoration both in the photographic and heterogeneous domains. 
\begin{figure}[h]
\begin{center}
   \includegraphics[width=7.5cm]{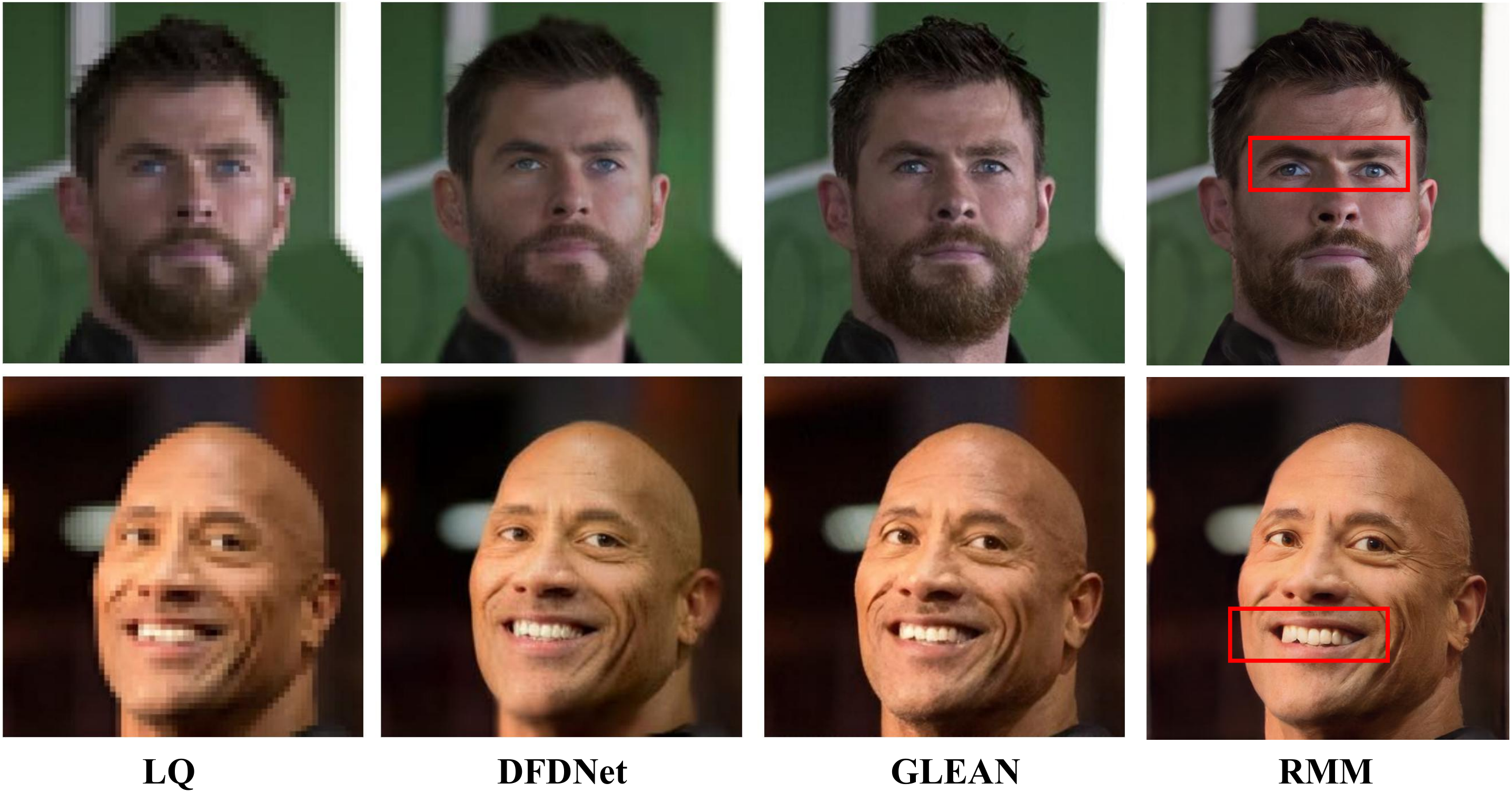}
\end{center}
\vspace{-13.pt}
   \caption{The comparison results of DFDNet, GLEAN and RMM. \textbf{Zoom in for the best view.}}
\label{fig:noise}
\end{figure}
\paragraph{Additional Analysis of RMM}
We set the dimension of $Z_{W}$ and $Z_{N}$ to 765 and 512, respectively. If randomly sampling different $Z_{N}$ or choosing different top-$i$ wavelet memory $Z_{W}$ to implement the modulation in the inference stage, the visual results maintain the same. However, the residual maps in Figure \ref{fig:ana} (up) demonstrate that the low-rank wavelet memory has more mistakes of the high-frequency coefficient, e.g., the eye area of $X_{SR}^{wm_{300}}-X_{SR}^{wm_{0}}$, while using the same $Z_{N}$.  Furthermore, we use the same top-1 wavelet memory but different noises as the universal prior, and show the results in Figure \ref{fig:ana} (below). We find that the global spatial information of the residual maps is maintained, although different noises focus on varied image areas for BFR.

\paragraph{Motivation of embedding prior} 
We think the key challenge of Blind face restoration is the restoration controllability and generalization. Our main aim is to restore the high-fidelity face with respect to the original structure and textual distribution, and improve the generalization in the wild and heterogeneous domains. To address this issue, we propose an effective and neat framework RMM. Specifically, the spatial feature $Z_{S}$ helps to preserve the scene content and face identity by maintaining the global topology information. The wavelet style code $Z_{W}$ helps to restore the high-frequency textual details by matching the memory knowledge. 
\begin{figure}[ht]
\begin{center}
   \includegraphics[width=8cm]{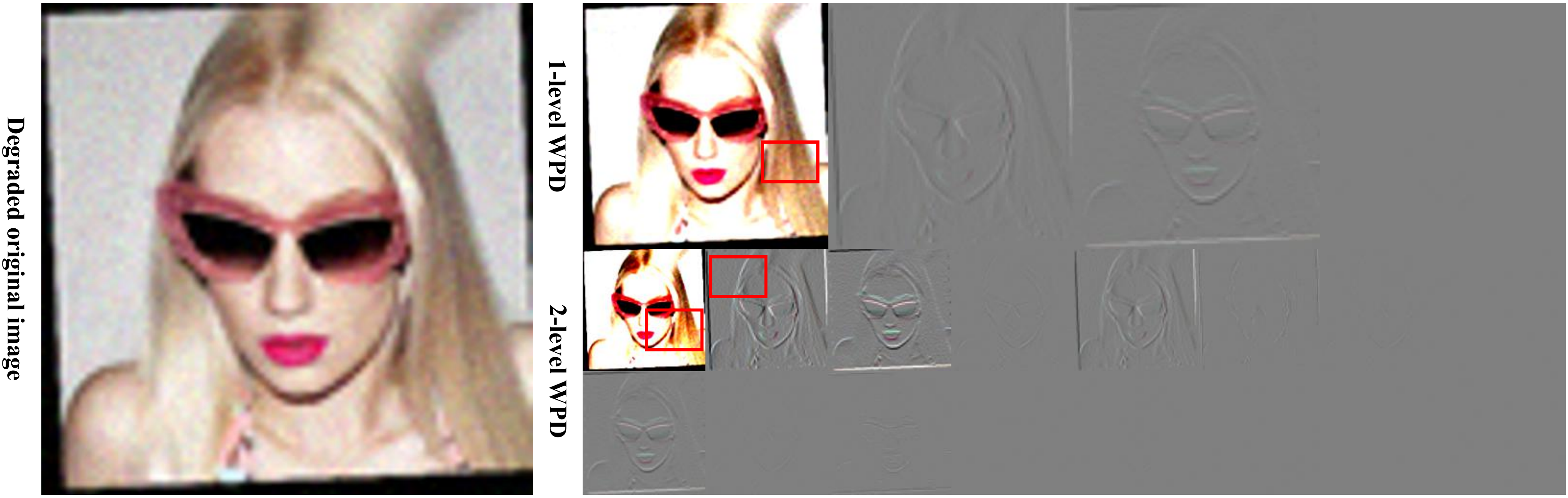}
\end{center}
\vspace{-13.pt}
   \caption{1-level and 2-level full wavelet packet decomposition (WPD) image. The random Gaussian noise in the original image remains in the WPD image as well. \textbf{Zoom in for the best view.}}
\label{fig:wpd}
\end{figure}
\begin{figure}[ht]
\begin{center}
   \includegraphics[width=8cm]{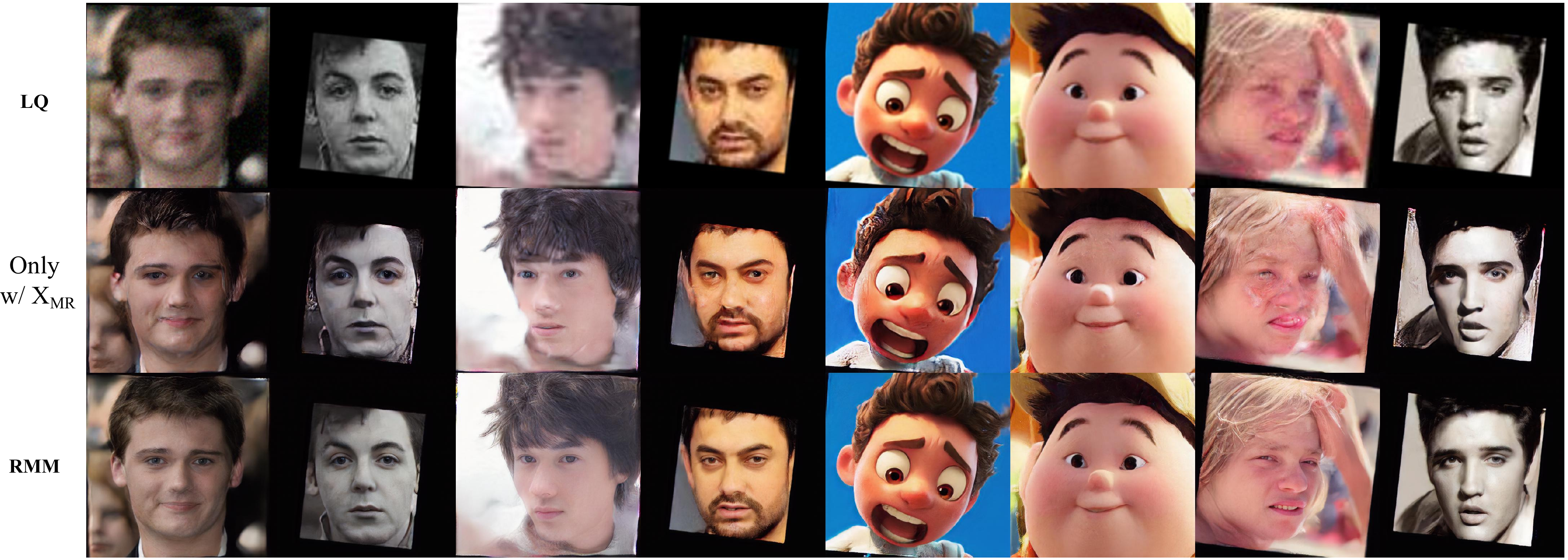}
\end{center}
\vspace{-13.pt}
   \caption{More examples only with $X_{MR}$. \textbf{Zoom in for the best view.}}
   \vspace{-13.pt}
\label{fig:noise}
\end{figure}
Note that wavelet packet decomposition (WPD) only considers the image magnification, and the random Gaussian noise in the original image remains in the WPD image as well, as shown in Figure \ref{fig:wpd}. Based on the strong representation of wavelet coefficients, WaveletSR[3] has studied the face super-resolution task. However, BFR task is more challenging than FSR task. In the real wild, the imaging and storage of image is easily interfered, which is very challenging for the model generalization. Therefore, the proposed universal prior $Z_{N}$ is used to improve the model robustness by defending additional blind degradation pattern, such as motion blur, Gaussian noise, JPEG compression.  Extensive experiments in the manuscript have shown that the proposed modulations of global spatial, universal prior and memorized wavelet embedding are beneficial and indispensable to restoring high-fidelity face in the complex scenes, and generalizing in the wild.

\begin{figure}[ht]
\begin{center}
   \includegraphics[width=8.2cm]{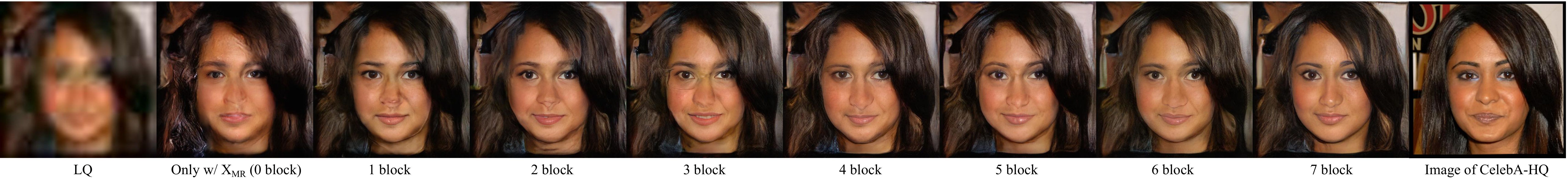}
\end{center}
\vspace{-13.pt}
   \caption{The results corresponding to RMM variants with different blocks. \textbf{Zoom in for the best view.}}
   \vspace{-13.pt}
\label{fig:blocks}
\end{figure}

\paragraph{RMM block size} We conduct the experiment considering the number of $RM^{3}$ blocks, and only with $X_{MR}$, i.e., $RM^{3}\_0$, as shown in Table \ref{tab:blk}. We apply three widely used metrics in the wild to evaluate the RMM variants on four test set, and obtain the mean scores. $RM^{3}\_7$ has the best scores, which indicates the progressive framework is effective, as shown in Figure \ref{fig:blocks}.
\begin{table}[h]
\centering
\caption{Quantitative evaluation on RMM variants with different $RM^{3}$ blocks.}
\resizebox{3.2cm}{1.45cm}{
\begin{tabular}{|c|ccc|}
\hline 
Dataset & \multicolumn{3}{c|}{Mean}\tabularnewline
\hline 
\makecell[c]{Methods} & \makecell[c]{FID$\downarrow$} & \makecell[c]{KID$\downarrow$} & NIQE$\downarrow$\tabularnewline
\hline 
$RM^{3}$\_0 & 111.46&	10.52&	4.7371\tabularnewline
$RM^{3}$\_1 & 104.16&	10.63&	4.5473\tabularnewline
$RM^{3}$\_2 & 107.51&	8.80&	4.5660\tabularnewline
$RM^{3}$\_3 & 103.10&	8.36&	4.6206\tabularnewline
$RM^{3}$\_4 & 106.20&	9.00&	4.5533\tabularnewline
$RM^{3}$\_5 & 101.29&	8.29&	4.5789\tabularnewline
$RM^{3}$\_6 & 100.81&	8.36&	4.5897\tabularnewline
$RM^{3}$\_7 & \textbf{100.61}&	\textbf{8.24}& \textbf{4.5415}\tabularnewline
\hline
\end{tabular}
}
\label{tab:blk}
\end{table}
\begin{figure}[h]
\begin{center}
   \includegraphics[width=4cm]{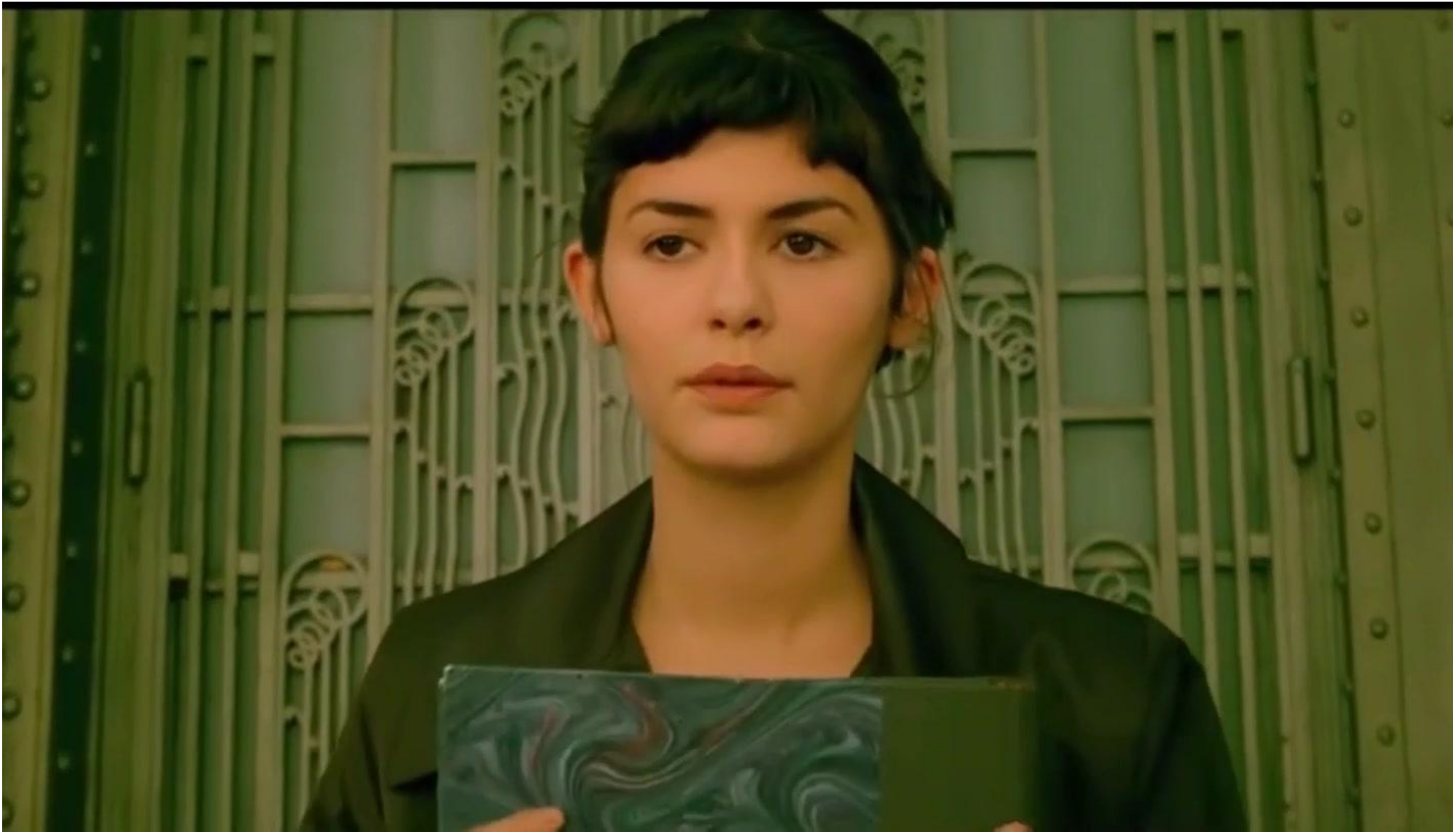}\includegraphics[width=4cm]{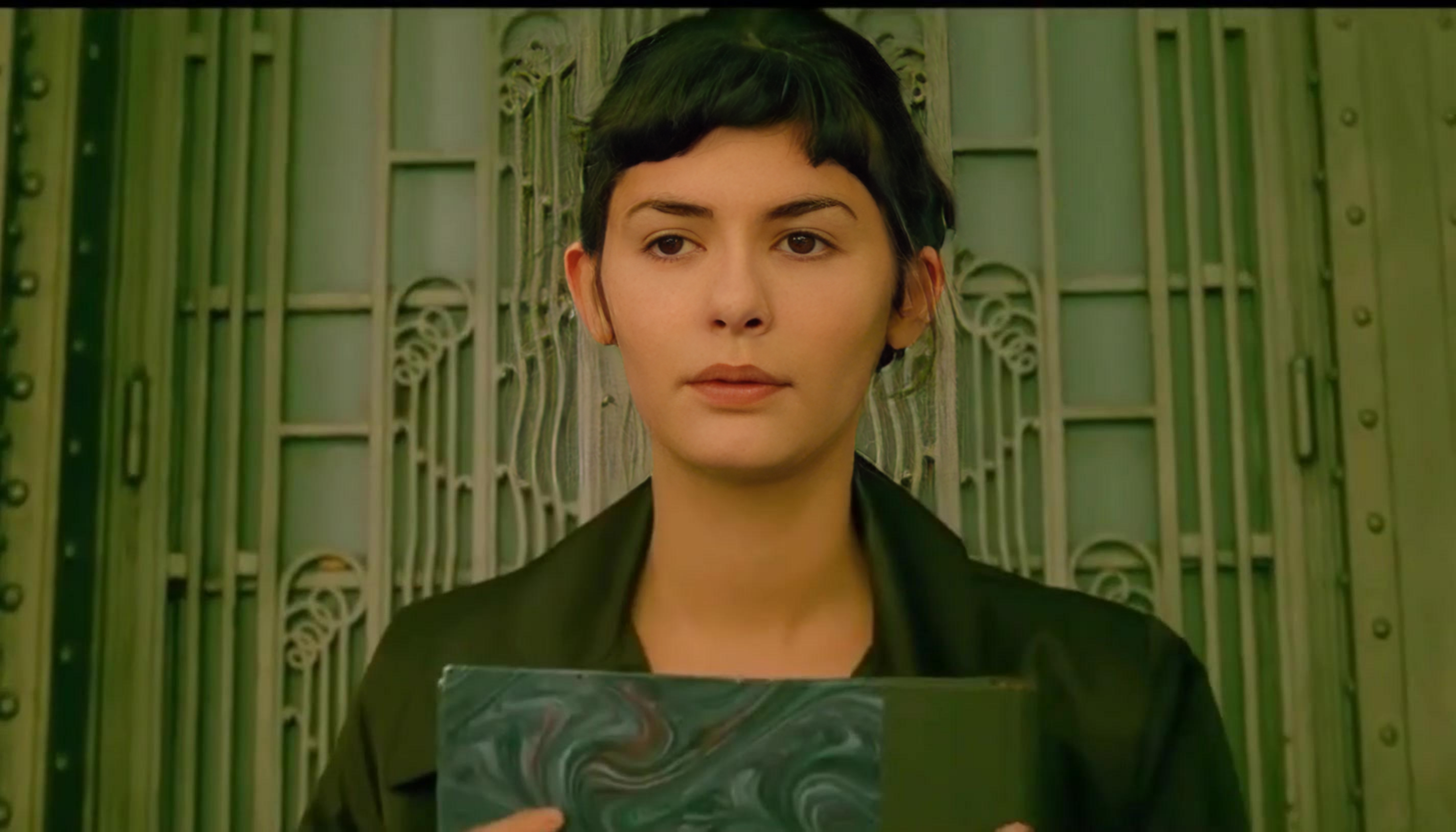}
\end{center}
\vspace{-13.pt}
   \caption{The right is the result of RMM applied to the movie $\textbf{Amelie}$ (2001). \textbf{Zoom in for the best view}. Particularly, the eyes and lower eyelashes of the right are higher-quality.}
\label{fig:ts}
\end{figure}
\begin{figure*}[h]
\begin{center}
   \includegraphics[width=17cm]{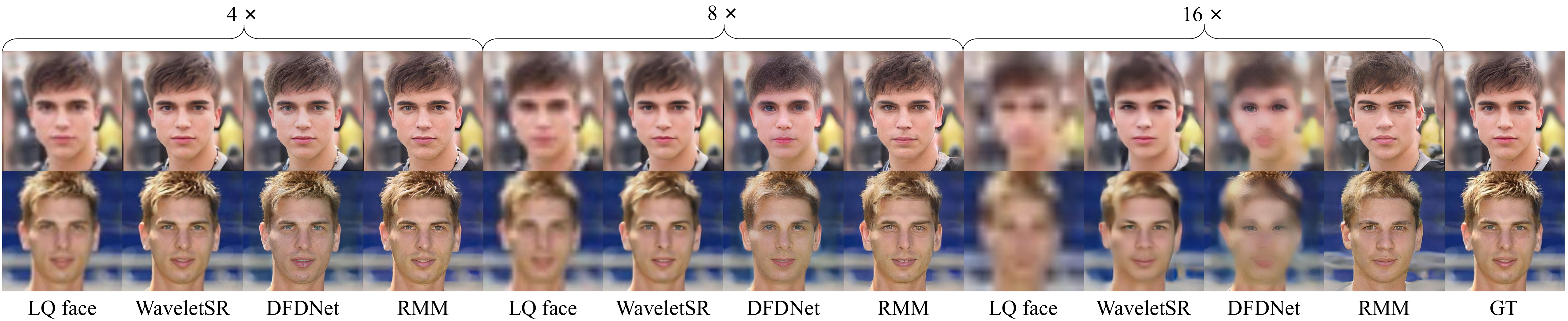}
\end{center}
\vspace{-13.pt}
   \caption{Qualitative comparison with WaveletSR \cite{wavelet2017} and DFDNet \cite{dfdnet2020} for FSR of different down-sampling scales on CelebA-TD (row 1) and VGGFace2-TD (row 2).}
\label{fig:sr}
\end{figure*}

\begin{table*}[h]
\caption{Quantitative evaluation on CelebA-TD and VGGFace2-TD. We train WaveletSR\cite{wavelet2017} based on FFHQ \cite{style2019} dataset, denoted as $WaveletSR^{*}$. The red and blue values mean the top-1 and top-2 scores. Note that Bicubic and GT are not involved in scoring.}
\resizebox{17.5cm}{1.95cm}{
\begin{tabular}{|c|ccccccccc|ccccccccc|}
\hline 
\makecell[c]{Methods} & \makecell[c]{LPIPS$\downarrow$} & \makecell[c]{FID$\downarrow$ \\(FFHQ/CelebA)} & \makecell[c]{KID$\times 100 \downarrow$\\(FFHQ/CelebA)} & \makecell[c]{NIQE$\downarrow$} & \makecell[c]{MS-SSIM$\uparrow$} & \makecell[c]{PSNR$\uparrow$} & \makecell[c]{SSIM$\uparrow$} & \makecell[c]{FED$\downarrow$} & LLE$\downarrow$ & \makecell[c]{LPIPS$\downarrow$} & \makecell[c]{FID$\downarrow$\\(FFHQ/VGGFace2)} & \makecell[c]{KID$\times 100 \downarrow$\\(FFHQ/VGGFace2)} & \makecell[c]{NIQE$\downarrow$} & \makecell[c]{MS-SSIM$\uparrow$} & \makecell[c]{PSNR$\uparrow$} & \makecell[c]{SSIM$\uparrow$} & \makecell[c]{FED$\downarrow$} & LLE$\downarrow$\tabularnewline
\hline 
 & \multicolumn{9}{c|}{CelebA 4$\times$} & \multicolumn{9}{c|}{VGGFace 4$\times$}\tabularnewline
\hline 
Bicubic & 0.33 & 177.17 / 53.14 & 17.99 / 5.80 & 10.05 & 0.94 & 25.05 & 0.78 & 0.28 & 1.31 & 0.17 & 119.37 / 17.48 & 11.82 / 1.30 & 10.41 & 0.97 & 31.73 & 0.88 & 0.20 & 1.22\tabularnewline
WaveletSR{*}\cite{wavelet2017} & 0.15 & 108.42 / \textbf{\textcolor{blue}{23.86}} & 10.66 / 1.89 & 6.95 & \textbf{\textcolor{red}{0.97}} & \textbf{\textcolor{red}{26.62}} & \textbf{\textcolor{blue}{0.75}} & \textbf{\textcolor{red}{0.17}} & \textbf{\textcolor{red}{0.98}} & \textbf{\textcolor{blue}{0.21}} & 82.63 / \textbf{\textcolor{red}{32.63}} & 7.26 / \textbf{\textcolor{red}{2.63}} & 8.01 & \textbf{\textcolor{red}{0.96}} & \textbf{\textcolor{red}{27.92}} & \textbf{\textcolor{red}{0.80}} & \textbf{\textcolor{red}{0.22}} & \textbf{\textcolor{blue}{1.58}}\tabularnewline
DFDNet\cite{dfdnet2020} & \textbf{\textcolor{blue}{0.11}} & \textbf{\textcolor{blue}{91.32}} / \textbf{\textcolor{red}{16.17}} & \textbf{\textcolor{blue}{7.69}} /\textbf{\textcolor{red}{{} 0.72}} & \textbf{\textcolor{red}{4.05}} & 0.92 & 25.31 & 0.74 & 0.22 & 1.38 & 0.23 & \textbf{\textcolor{blue}{54.60 / 42.79}} & \textbf{\textcolor{blue}{3.82 / 3.55}} & \textbf{\textcolor{red}{4.22}} & 0.91 & 25.46 & 0.75 & 0.26 & 1.82\tabularnewline
RMM & \textbf{\textcolor{red}{0.09}} & \textbf{\textcolor{red}{77.32}} / 27.28 & \textbf{\textcolor{red}{4.59}} / \textbf{\textcolor{blue}{1.28}} & \textbf{\textcolor{blue}{4.26}} & \textbf{\textcolor{blue}{0.95}} & \textbf{\textcolor{blue}{26.47}} & \textbf{\textcolor{red}{0.77}} & \textbf{\textcolor{blue}{0.19}} & \textbf{\textcolor{blue}{1.10}} & \textbf{\textcolor{red}{0.20}} & \textbf{\textcolor{red}{46.19}} / 64.49 & \textbf{\textcolor{red}{2.27}} / 5.58 & \textbf{\textcolor{blue}{4.35}} & \textbf{\textcolor{blue}{0.94}} & \textbf{\textcolor{blue}{27.47}} & \textbf{\textcolor{blue}{0.78}} & \textbf{\textcolor{blue}{0.24}} & \textbf{\textcolor{red}{1.57}}\tabularnewline
\hline 
 & \multicolumn{9}{c|}{CelebA 8$\times$} & \multicolumn{9}{c|}{VGGFace 8$\times$}\tabularnewline
\hline 
Bicubic & 0.53 & 160.76 / 68.04 & 14.71 / 6.11 & 11.53 & 0.84 & 21.23 & 0.65 & 0.55 & 2.64 & 0.36 & 134.35 / 34.39 & 13.16 / 2.44 & 11.67 & 0.89 & 25.33 & 0.76 & 0.47 & 2.59\tabularnewline
WaveletSR{*}\cite{wavelet2017} & 0.25 & 123.70 / 39.71 & 11.01 / 3.18 & 7.51 & \textbf{\textcolor{red}{0.93}} & \textbf{\textcolor{red}{23.51}} & \textbf{\textcolor{blue}{0.64}} & \textbf{\textcolor{blue}{0.46}} & \textbf{\textcolor{red}{1.88}} & \textbf{\textcolor{blue}{0.26}} & 78.51 / \textbf{\textcolor{red}{30.37}} & 6.25 / \textbf{\textcolor{red}{2.03}} & 8.01 & 0.83 & \textbf{\textcolor{red}{24.56}} & 0.70 & \textbf{\textcolor{blue}{0.46}} & 2.49\tabularnewline
DFDNet\cite{dfdnet2020} & \textbf{\textcolor{blue}{0.21}} & \textbf{\textcolor{blue}{106.21}} / \textbf{\textcolor{red}{31.83}} & \textbf{\textcolor{blue}{8.99 }}/ \textbf{\textcolor{red}{2.03}} & \textbf{\textcolor{red}{4.09}} & 0.86 & \textbf{\textcolor{blue}{23.28}} & \textbf{\textcolor{red}{0.65}} & 0.50 & 1.97 & 0.29 & \textbf{\textcolor{blue}{57.55 / 55.97}} & \textbf{\textcolor{blue}{3.97 / 4.93}} & \textbf{\textcolor{blue}{4.38}} & \textbf{\textcolor{blue}{0.86}} & \textbf{\textcolor{blue}{24.38}} & \textbf{\textcolor{red}{0.70}} & 0.48 & \textbf{\textcolor{blue}{2.40}}\tabularnewline
RMM & \textbf{\textcolor{red}{0.17}} & \textbf{\textcolor{red}{71.97}} / \textbf{\textcolor{blue}{38.48}} & \textbf{\textcolor{red}{3.76}} / \textbf{\textcolor{blue}{2.27}} & \textbf{\textcolor{blue}{4.15}} & \textbf{\textcolor{blue}{0.87}} & 22.15 & 0.63 & \textbf{\textcolor{red}{0.45}} & \textbf{\textcolor{blue}{1.91}} & \textbf{\textcolor{red}{0.21}} & \textbf{\textcolor{red}{45.92}} / 68.59 & \textbf{\textcolor{red}{2.25}} / 5.76 & \textbf{\textcolor{red}{4.37}} & \textbf{\textcolor{red}{0.87}} & 24.29 & \textbf{\textcolor{blue}{0.67}} & \textbf{\textcolor{red}{0.44}} & \textbf{\textcolor{red}{2.32}}\tabularnewline
\hline 
 & \multicolumn{9}{c|}{CelebA 16$\times$} & \multicolumn{9}{c|}{VGGFace 16$\times$}\tabularnewline
\hline 
Bicubic & 0.66 & 187.05 / 158.83 & 16.79 / 15.04 & 12.69 & 0.66 & 17.88 & 0.54 & 0.71 & 6.55 & 0.53 & 184.38 / 114.99 & 16.69 / 9.83 & 12.63 & 0.71 & 20.75 & 0.64 & 0.65 & 6.21\tabularnewline
WaveletSR{*}\cite{wavelet2017} & \textbf{\textcolor{blue}{0.34}} & \textbf{\textcolor{blue}{120.27}} / \textbf{\textcolor{red}{62.27}} & \textbf{\textcolor{blue}{10.74}} / \textbf{\textcolor{red}{5.01}} & 6.04 & \textbf{\textcolor{blue}{0.72}} & \textbf{\textcolor{blue}{19.42}} & \textbf{\textcolor{blue}{0.61}} & \textbf{\textcolor{blue}{0.59}} & \textbf{\textcolor{blue}{2.35}} & \textbf{\textcolor{blue}{0.31}} & 95.89 / \textbf{\textcolor{red}{49.45}} & 8.35 / \textbf{\textcolor{red}{3.59}} & 6.21 & \textbf{\textcolor{blue}{0.78}} & \textbf{\textcolor{blue}{22.33}} & 0.60 & \textbf{\textcolor{blue}{0.58}} & \textbf{\textcolor{blue}{3.84}}\tabularnewline
DFDNet\cite{dfdnet2020} & 0.40 & 143.18 / 80.31 & 12.94 / 6.85 & \textbf{\textcolor{blue}{5.16}} & 0.70 & 19.03 & 0.53 & 0.66 & 4.24 & 0.39 & \textbf{\textcolor{blue}{88.59}} / 87.93 & \textbf{\textcolor{blue}{7.02}} / 8.40 & \textbf{\textcolor{blue}{5.54}} & 0.75 & 22.17 & \textbf{\textcolor{blue}{0.66}} & 0.64 & 4.57\tabularnewline
RMM & \textbf{\textcolor{red}{0.23}} & \textbf{\textcolor{red}{74.00}} / \textbf{\textcolor{blue}{73.75}} & \textbf{\textcolor{red}{4.25}} / \textbf{\textcolor{blue}{5.58}} & \textbf{\textcolor{red}{4.43}} & \textbf{\textcolor{red}{0.78}} & \textbf{\textcolor{red}{20.39}} & \textbf{\textcolor{red}{0.66}} & \textbf{\textcolor{red}{0.58}} & \textbf{\textcolor{red}{2.30}} & \textbf{\textcolor{red}{0.29}} & \textbf{\textcolor{red}{55.04}} /\textbf{\textcolor{blue}{{} 83.44}} & \textbf{\textcolor{red}{2.96}} / \textbf{\textcolor{blue}{6.87}} & \textbf{\textcolor{red}{4.54}} & \textbf{\textcolor{red}{0.83}} & \textbf{\textcolor{red}{23.87}} & \textbf{\textcolor{red}{0.67}} & \textbf{\textcolor{red}{0.57}} & \textbf{\textcolor{red}{3.12}}\tabularnewline
\hline 
GT & \makecell[c]{0} & \makecell[c]{87.97 / 0} & \makecell[c]{6.99 / 0} & \makecell[c]{4.62} & \makecell[c]{1} & \makecell[c]{$\infty$} & \makecell[c]{1} & \makecell[c]{0} & 0 & \makecell[c]{0} & \makecell[c]{96.16 / 0} & \makecell[c]{8.28 / 0} & \makecell[c]{6.98} & \makecell[c]{1} & \makecell[c]{$\infty$} & \makecell[c]{1} & \makecell[c]{0} & 0\tabularnewline
\hline 
\end{tabular}}
\label{tab: one}
\end{table*}
\subsection*{Face Super Resolution}
We compare our RMM with WaveletSR \cite{wavelet2017} and DFDNet \cite{dfdnet2020} on the CelebA-TD and VGGFace2-TD, as shown in Table \ref{tab: one}. The perceptual fidelity of FFHQ is higher than \cite{celeb2017, vggface2_2018}, and we get better quantitative results, e.g., FID, KID, compared with other state-of-the-art methods. DFDNet \cite{dfdnet2020} is not competent to dealing with the LQ images with 16 $\times$ downsampling degradation, as shown in Figure \ref{fig:sr}. Note that WaveletSR \cite{wavelet2017} predicts high-frequency wavelet coefficients, so the pixel-wise perception and identity preservation metrics are better in 4 $\times$ inference setting. However, the image fidelity of WaveletSR \cite{wavelet2017}, e.g., FID, KID, or NIQE, is not competitive. Note that the original test image of VGGFace2 is usually not high-quality, whereas our results are high-fidelity, so FID-VGGFace2 of our RMM is higher. More details are shown in Table \ref{tab: one}.

\begin{figure*}[h]
\begin{center}
   \includegraphics[width=16.2cm]{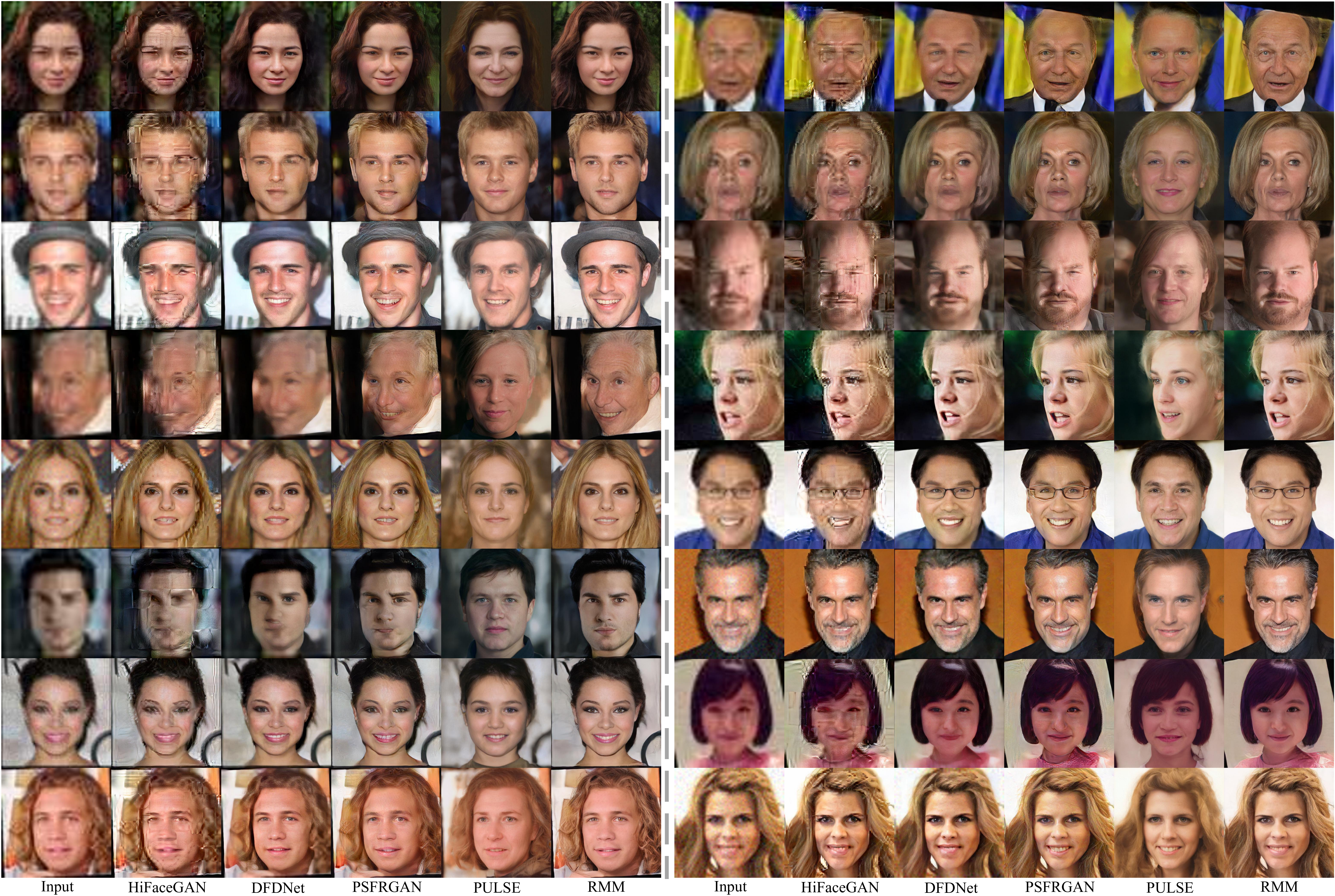}
   
   \includegraphics[width=16.2cm]{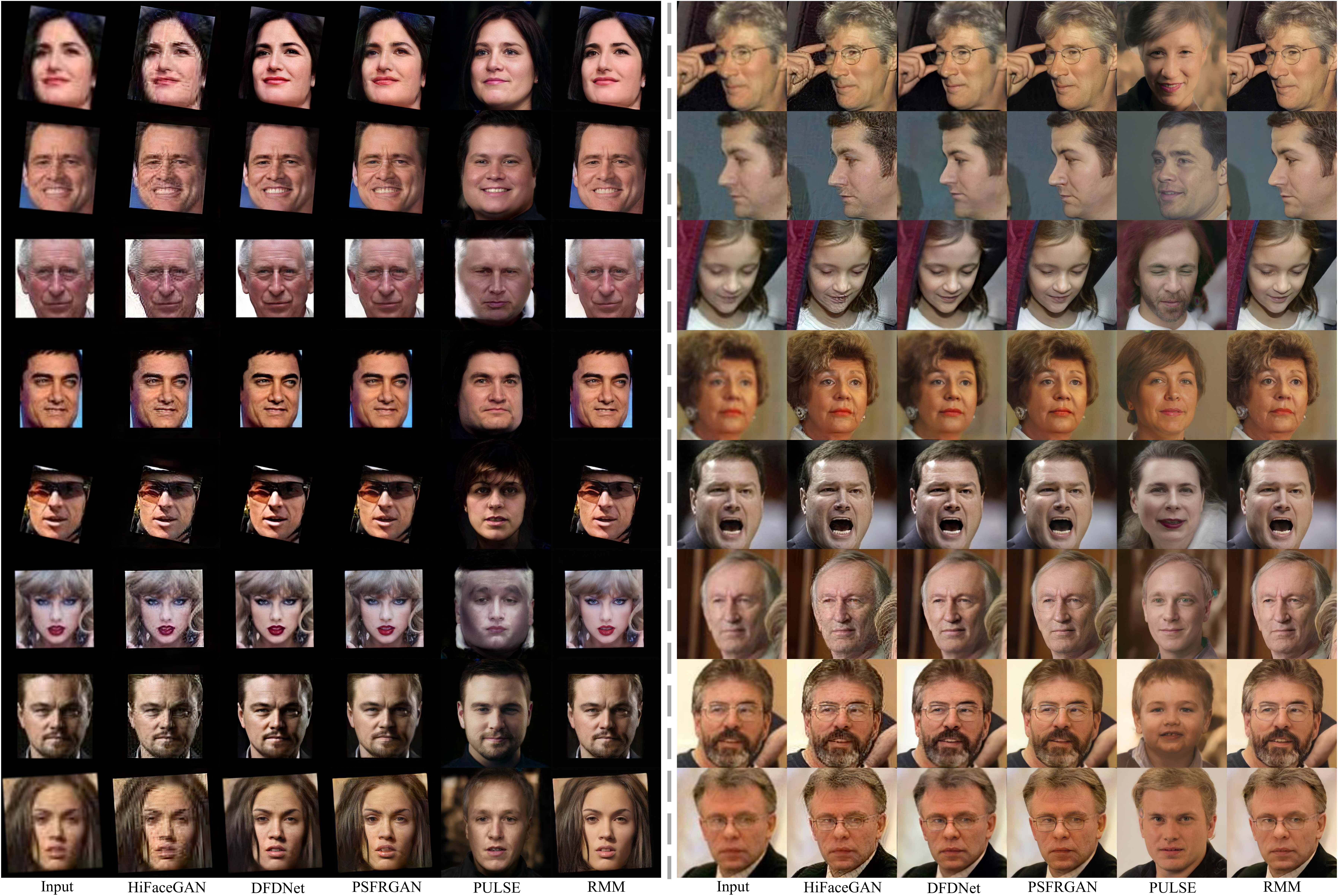}
\end{center}
\vspace{-13.pt}
   \caption{More results on CelebA-TB (left-up), VGGFace2-TB (right-up), CFW-Test (left-below) and LFW-Test (right-below). \textbf{Zoom in for the best view.}}
\label{fig:celeb-vgg}
\end{figure*}

\paragraph{Qualitative Results of RMM} We show more results in Figure \ref{fig:celeb-vgg} on CelebA-TB, VGGFace2-TB, CFW-Test and LFW-Test, respectively. 
Moreover, we provide a demo of RMM in the supplementary material, and the video clip is from $\textbf{Amelie}$ (2001). The original resolution is set to 1424 $\times$ 814, and the restored frame is 2848 $\times$ 1628. We show the comparison of a frame in Figure \ref{fig:ts}. For the stability of the background, we paste the restored face to the original frame using an open-source face parsing model. 

\end{document}